%% file: main.tex
\documentclass[10pt,journal,compsoc]{IEEEtran}
\usepackage{color}
\usepackage{bm}
\usepackage{booktabs}
\usepackage{adjustbox}
\usepackage{diagbox}
\usepackage{multirow}
\usepackage{comment}
\usepackage[normalem]{ulem}
\usepackage{graphicx}
\usepackage{tikz}
\usepackage{amsmath,amssymb} 
\usepackage[ruled]{algorithm2e}
\usepackage{subfigure}
\usepackage{lipsum}
\usepackage{enumitem}
\usepackage{multicol}
\usepackage{tabularx,ragged2e}
\usepackage{color, colortbl}
\usepackage{wrapfig}

\def\ie{\emph{i.e.}}
\def\eg{\emph{e.g.}}

\def\etal{\emph{et al.}}

\newcommand{\bH}{{\mathbf{H}}}
\newcommand{\bJ}{{\mathbf{J}}}
\newcommand{\hbH}{{\hat{\mathbf{H}}}}
\newcommand{\hbJ}{{\hat{\mathbf{J}}}}
\newcommand{\tbH}{{\tilde{\mathbf{H}}}}

\newcommand{\bT}{{\mathbf{T}}}
\newcommand{\bp}{{\mathbf{p}}}
\newcommand{\bq}{{\mathbf{q}}}
\newcommand{\bw}{{\mathbf{w}}}
\newcommand{\bmu}{{\bm{\mu}}}

\definecolor{darkcyan}{rgb}{0.0, 0.55, 0.55}
\definecolor{olivine}{rgb}{0.6, 0.73, 0.45}
\definecolor{pastelred}{rgb}{1.0, 0.41, 0.38}

\makeatletter
\long\def\@makecaption#1#2{%
  \vskip\abovecaptionskip
  \sbox\@tempboxa{#1: #2}%
  \ifdim \wd\@tempboxa >\hsize
    #1: #2\par
  \else
    \global \@minipagefalse
    \hb@xt@\hsize{\box\@tempboxa\hfil}%
  \fi
  \vskip\belowcaptionskip}
\makeatother

\ifCLASSOPTIONcompsoc
\else
  \usepackage{cite}
\fi
\ifCLASSINFOpdf
\else
\fi
\begin{document}
%
\title{Bias-Compensated Integral Regression \\
for Human Pose Estimation}
%
%
%
%

\author{Kerui~Gu,
        Linlin~Yang,
        Michael~Bi~Mi,
        Angela~Yao
\IEEEcompsocitemizethanks{\IEEEcompsocthanksitem K. Gu, L. Yang, and A. Yao are with the 
School of Computing at the National University of Singapore. 
E-mail: \{keruigu, yangll, ayao\}@comp.nus.edu.sg.
\IEEEcompsocthanksitem M. Bi is with Huawei International Pte Ltd.}
}

%
%

\markboth{Journal of \LaTeX\ Class Files,~Vol.~14, No.~8, August~2015}%
{Shell \MakeLowercase{\textit{et al.}}: Bare Demo of IEEEtran.cls for Computer Society Journals}
%



\IEEEtitleabstractindextext{%
\begin{abstract}
   In human and hand pose estimation, heatmaps are a crucial intermediate representation for a body or hand keypoint. Two popular methods to decode the heatmap into a final joint coordinate are via an argmax, as done in heatmap detection, or via softmax and expectation, as done in integral regression. 
   Integral regression is learnable end-to-end, but has lower accuracy than detection. This paper uncovers an induced bias from integral regression that results from combining the softmax and the expectation operation. This bias often forces the network to learn degenerately localized heatmaps, obscuring the keypoint's true underlying distribution and leads to lower accuracies. Training-wise, by investigating the gradients of integral regression, we show that the implicit guidance of integral regression to update the heatmap makes it slower to converge than detection. To counter the above two limitations, we propose Bias Compensated Integral Regression (BCIR), an integral regression-based framework that compensates for the bias.  BCIR also incorporates a Gaussian prior loss to speed up training and improve prediction accuracy. Experimental results on both the human body and hand benchmarks show that BCIR is faster to train and more accurate than the original integral regression, making it competitive with state-of-the-art detection methods.
\end{abstract}

\begin{IEEEkeywords}
Human Pose Estimation, Hand Pose Estimation, Latent Regression, Numerical Coordinate Regression, Heatmap Regression, Heatmap Decoding, Bias Compensation
\end{IEEEkeywords}}

\maketitle

\IEEEdisplaynontitleabstractindextext

%
\IEEEpeerreviewmaketitle

\newpage
\newpage

\section{Introduction}
In computer vision, 2D (human) pose estimation aims to determine the spatial positions of articulated joints of the full body or the hand. At first glance, pose estimation seems to be a straightforward regression problem. However, methods that directly regress joint coordinates~\cite{carreira2016human, sun2018integral, nibali2018numerical, li2021pose} are less effective than those which locate joints by estimating a likelihood heatmap~\cite{newell2016stacked, chen2018cascaded, xiao2018simple, sun2019deep,sun2018integral,iqbal2018hand}. The rationale is that working with heatmaps allows neural networks to keep a fully convolutional architecture and therefore retain critical spatial structures throughout the encoding and decoding process~\cite{tompson2014joint, zhang2020distribution}.

When estimating a likelihood heatmap, a deep neural network is tasked with predicting, at each pixel, the probability of that pixel being a given joint. A common method to decode a heatmap into a joint coordinate is to formulate it as a maximum likelihood problem and use an $\emph{argmax}$ function~\cite{newell2016stacked, chen2018cascaded, xiao2018simple, sun2019deep}. In some ways, this form of decoding is analogous to the frameworks used in object detection, and such an approach has been referred to as~\emph{detection-based}~\cite{sun2018integral}.  \footnote{Our work will also use this naming convention.  We raise the reader's awareness of possible confusion with other naming conventions used in literature, which sometimes refers to detection-based methods as indirect regression or heatmap regression~\cite{bulat2016human, iqbal2018hand, zhang2020distribution}}.

An alternative form of decoding is to take a $\emph{softmax}$ together with an expectation~\cite{iqbal2018hand,sun2018integral,nibali2018numerical,luvizon2019human}. Unlike the argmax, both the softmax and expectation function are differentiable. As such, one can learn directly from ground truth joint coordinates while retaining the benefits of fully convolutional network architectures.  This approach is referred to as \emph{latent heatmap regression}~\cite{iqbal2018hand} in the hand pose estimation literature and \emph{integral pose estimation (IPR)}~\cite{sun2018integral} in the body pose estimation literature.  Fig.~\ref{fig:overview} shows a  comparison of the pipeline in detection and integral regression methods.  

In principle, integral regression should outperform detection methods for pose estimation. Firstly, they are {end-to-end trainable} without imposing a (Gaussian) structure on the heatmap. Secondly, decoding with an expectation operation does not fix the estimated coordinate to the (discrete) resolution of the heatmap. The comparison between integral regression and detection methods from Sun~\etal~\cite{sun2018integral} concluded that integral regression is either as competitive as or better than detection for 2D pose estimation. However, the results are not fully conclusive because~\cite{sun2018integral} used a unique backbone for the integral regression pipeline even though the image encoding should be the same for both methods (see Fig.~\ref{fig:overview}).  

Given that detection methods are still the predominant state-of-the-art techniques for human pose estimation~\cite{newell2016stacked, chen2018cascaded, xiao2018simple, sun2019deep}, this paper aims to investigate the differences between detection and integral regression to explain why, in practice, integral regression tends to lag in performance. Based on the heatmap decoding methods and supervision, we elaborate on possible causes of the performance differences  with theoretical support. To begin with, we highlight two key experimental findings. 

\textbf{Finding 1: Integral regression outperforms detection for ``hard'' cases of human pose estimation, while detection methods excel at ``easy'' cases.}  Splitting the pose estimation benchmarks 
according to factors like the number of joints and occlusion reveals an {uneven} performance advantage between the two methods (see split details in Sec.~\ref{factors}).  Detection excels and outperforms integral regression on ``easy'' test samples, which have most or all keypoints present in the scene, little or no occlusion of the keypoints, and higher bounding box resolution.  However, for ``hard'' cases, \ie, scenes with fewer keypoints, higher occlusion, and lower resolutions, integral regression consistently outperforms detection. These effects have been previously obscured by the standard evaluation measure of a single mAP because the hard cases constitute the dataset tail. 

\textbf{Finding 2: Integral regression networks are slower to converge than detection methods}. Integral regression networks need around six times the number of epochs of detection to reach 80\% of their final accuracy. The difference in training speed has not been discussed explicitly in previous works.  

Theoretically, we can show that the expected end-point error (EPE) of soft-argmax decoding used in integral regression is always equal or lower than argmax decoding in detection.  Yet this result directly contradicts Finding 1. Our hypothesis is that the differences in the heatmap density are the underlying cause of this result.  Experimentally, we show that the heatmap spread of both methods shrinks when samples transition from ``hard'' to ``easy''.  Interestingly, integral regression shrinks at a faster rate to a very concentrated density. Furthermore, we show theoretically and experimentally that degenerately concentrated heatmap densities lead to {low pose accuracy}. Such heatmaps are more likely to be observed with integral regression methods on ``easy'' samples, hence the lower accuracy. 
This problem can be mitigated either by adding a regularizer or a spatial prior on the heatmap. 

Why does integral regression present extremely localized heatmaps? We conduct a deep dive into both the forward and back-propagation process. In the forward pass, we discover an induced bias in the decoding process. Specifically, combining the softmax with an expectation operation shifts the alignment of the heatmap with respect to the true coordinate position. More dispersed heatmap densities exacerbate this bias, \ie, the greater the heatmap spread, the larger the shift. To decode the predicted heatmaps and obtain the correct joint position, the network must implicitly learn to compensate for this shift, or as we have observed, produce more localized heatmaps and predict extremely large values on only a few pixels around the joint position. To counter these degenerate heatmaps and eliminate this bias, we propose the addition of a simple offset term to the decoding process. This simple compensation scheme is plug-and-play for all integral regression methods and significantly improves accuracy.

For the backward pass, we explore the update behavior of integral regression by explicitly deriving the gradients. In the initial stage of training, the gradients should give a large response around the ground truth location; however, for integral regression, the gradients tend to respond on pixels having large predicted values or heatmap, which do not usually overlap with the ground truth location.
Integral regression takes more epochs than detection to locate the ground truth in the heatmap, hence the slow convergence rate as stated in Finding 2. A dense form of supervision, \eg, the aforementioned spatial prior, can give indications of the ground truth location in the heatmap to speed up training convergence.

In summary, this paper presents a localized heatmap model that unifies the performance behavior of both integral regression and detection methods. By merging our theoretical and empirical findings, we arrive at a bias-compensated integral regression (BCIR) method.  Together with either a regularizer or a spatial prior, BCIR is fast to train and has competitive accuracy on both human and hand pose estimation benchmarks. 

Preliminary versions of this paper appeared in~\cite{gu2021removing} and~\cite{gu2022dive}.  The work of~\cite{gu2021removing} highlighted the accuracy differences between integral regression and detection and exposed the bias of integral regression.  The follow-up work in~\cite{gu2022dive} proposed the localized heatmap model and analyzed the backward pass of the integral pose regression method. The current paper merges the two works into a joint framework and offers a more comprehensive theoretical and empirical analysis of the localized heatmap model. Additionally, it links the heatmap collapse to the bias and extends the comparative experiments of integral and detection to hand pose estimation.

\section{Related Work}
Traditionally, human pose estimation used probabilistic models~\cite{pishchulin2013poselet, dantone2014body} or pictorial structures~\cite{felzenszwalb2005pictorial, andriluka2009pictorial} to capture the mutual relationship between the keypoints. Recent methods have become deep learning-based and use a convolutional neural network to directly map the input image to joint location predictions.

Early benchmarks feature a single person in the scene~\cite{modec13,Johnson10} but since the introduction of the MSCOCO benchmark~\cite{lin2014microsoft}, most works tackle multi-person 2D human pose estimation. Generally, methods can be divided into top-down~\cite{newell2016stacked, xiao2018simple, li2019rethinking, sun2019deep} or bottom-up approach~\cite{newell2016associative, fang2017rmpe, papandreou2018personlab, cao2019openpose, cheng2020higherhrnet, luo2021rethinking}. The top-down approach first applies a person detector to crop the human instances, followed by a single-person pose estimation method. Hence, the main aim of top-down methods is to generate accurate joint locations or heatmaps given an image containing only one human. Bottom-up methods first detect all possible keypoints in the scene and then group them into individuals. {Recent bottom-up methods mainly focus on designing association algorithms and usually adopt off-the-shelf backbones to detect all possible joints.}
Compared with bottom-up methods, top-down methods are less efficient but have consistently better {accuracy}. As our paper studies heatmap behavior, we focus on the setting of top-down methods.

\subsection{Regression-Based Methods}

Numerical regression-based methods directly regress the joint coordinates without the use of any heatmaps. The first deep learning-based approach, DeepPose~\cite{toshev2014deeppose}, followed this fashion and applied a series of convolution and fully connected layers to regress the coordinates. Numerical regression methods are commonly used in facial landmark detection~\cite{feng2018wing,zhu2020fast} but not human pose estimation.  Their accuracy is lower than heatmap-based methods likely because they capture less spatial knowledge~\cite{toshev2014deeppose, carreira2016human}. A recent transformer-based work~\cite{li2021pose} developed an encoder-decoder architecture in vision transformers~\cite{carion2020end, zhu2020deformable} to regress the joint coordinates but still cannot match the accuracy of state-of-the-art heatmap-based methods~\cite{cocoleaderboard}.

Instead of direct numerical regression, integral pose regression~\cite{levine2016end, yi2016lift, thewlis2017unsupervised,sun2018integral} uses a fully convolutional architecture while still regressing the joint coordinates. The ``integral'' takes the form of a soft-argmax with an expectation. While there are a large number of detection-based methods, only two recent works~\cite{luvizon2019human, nibali2018numerical} followed an integral regression approach. ~\cite{luvizon2019human} combined contextual information with a regression loss, while ~\cite{nibali2018numerical} proposed a variance or distribution penalty for integral regression.

Most works on hand pose estimation focused on 3D hand pose estimation. Early works either directly predicted 3D hand coordinates~\cite{zimmermann2017learning,yang2018disentangling,spurr2018cross,yang2019aligning,yu2021local} or lifted poses from 2D to 3D with networks trained with 2D heatmaps as intermediate supervision~\cite{cai2018weakly,zhang2019end}. To build explicit connections between the 2D and 3D space while maintaining a differentiable pipeline, subsequent works~\cite{iqbal2018hand,yang2021semihand,spurr2020weakly} proposed ``2.5D regression'' and integral regression with relative depth estimates.  Specifically, ~\cite{iqbal2018hand} first proposed the 2.5D regression and showed its ability to learn the arbitrary shape of each joint. \cite{spurr2020weakly} and~\cite{yang2021semihand} then applied 2.5D regression to the weakly- and semi-supervised setting, respectively.

\subsection{Detection-based Methods}
Since the heatmap representation was introduced in~\cite{tompson2014joint}, heatmap-based detection methods~\cite{gkioxari2016chained, wei2016convolutional, lifshitz2016human, chen2014articulated, chu2016crf, yang2016end, yang2017learning, liu2017active, chen2017adversarial} have dominated human pose estimation. A well-known example is the Hourglass Network~\cite{newell2016stacked}.  It stacks together encoder-decoder modules with skip connections to estimate and gradually refine the joint heatmaps.  The more recent Simple Baseline (SBL) architecture~\cite{xiao2018simple} combines a ResNet~\cite{he2016deep} and a few deconvolutional layers. In the High-Resolution Net (HRNet) work~\cite{sun2019deep}, the authors built a framework with denser skip connections and used higher resolutions for feature learning. Owing to their simple design and strong learning capabilities, SBL and HRNet have become standard backbones for other extensions. For example,~\cite{cai2020learning} leveraged these backbones to obtain dedicated local representations based on low-level spatial information, while  \cite{khirodkar2021multi} introduced a multi-instance modulation block to adaptively modulate channel-wise features produced by the two backbones.

\subsection{Heatmap Investigations}
Two recent works, UDP~\cite{huang2020devil} and DARK~\cite{zhang2020distribution}, have focused specifically on the decoding process of detection methods. 
UDP revealed a {data processing bias}
caused by the inconsistency in the coordinate system transformations during data augmentation.  {To reduce the bias, \cite{huang2020devil} proposed a unified coordinate system for different augmentations during data processing}.
Moreover, as the predicted heatmaps during inference may violate the Gaussian assumption and {make the location of the maximum value inaccurate}, 
UDP proposed to find the optimal Gaussian location based on the prediction and DARK directly modulated the predicted Gaussian distribution for decoding. Both UDP and DARK focused on the explicit heatmaps of detection-based methods and investigated the {heatmap distribution bias} under the Gaussian assumption.
However, {their proposed Gaussian unbiased training is not applicable for integral regression methods, which do not use the Gaussian assumption}. 

Only one work to date has closely investigated the heatmap in integral regression methods.~\cite{nibali2018numerical} experimentally compared different loss functions, heatmap normalization schemes, and regularizers. They found that using a Jensen-Shannon regularization on the heatmap with a softmax normalization and L1 loss achieved the best performance for integral regression. However, there still exists a significant performance gap between detection and integral methods. {Compared with~\cite{nibali2018numerical}, which simply presented an experimental study of regularization on heatmaps of integral regression, our work provides a comprehensive empirical and theoretical analysis of the differences between detection and integral methods and reveals an underlying drawback of integral regression, \ie, a bias in combining softmax with expectation. After removing the bias, we show that integral regression achieves superior performance compared with detection-based methods.}

\section{Preliminaries on Human Pose Estimation}\label{sec:prelim}

\begin{figure*}
    \centering
    \includegraphics[width=0.9\textwidth]{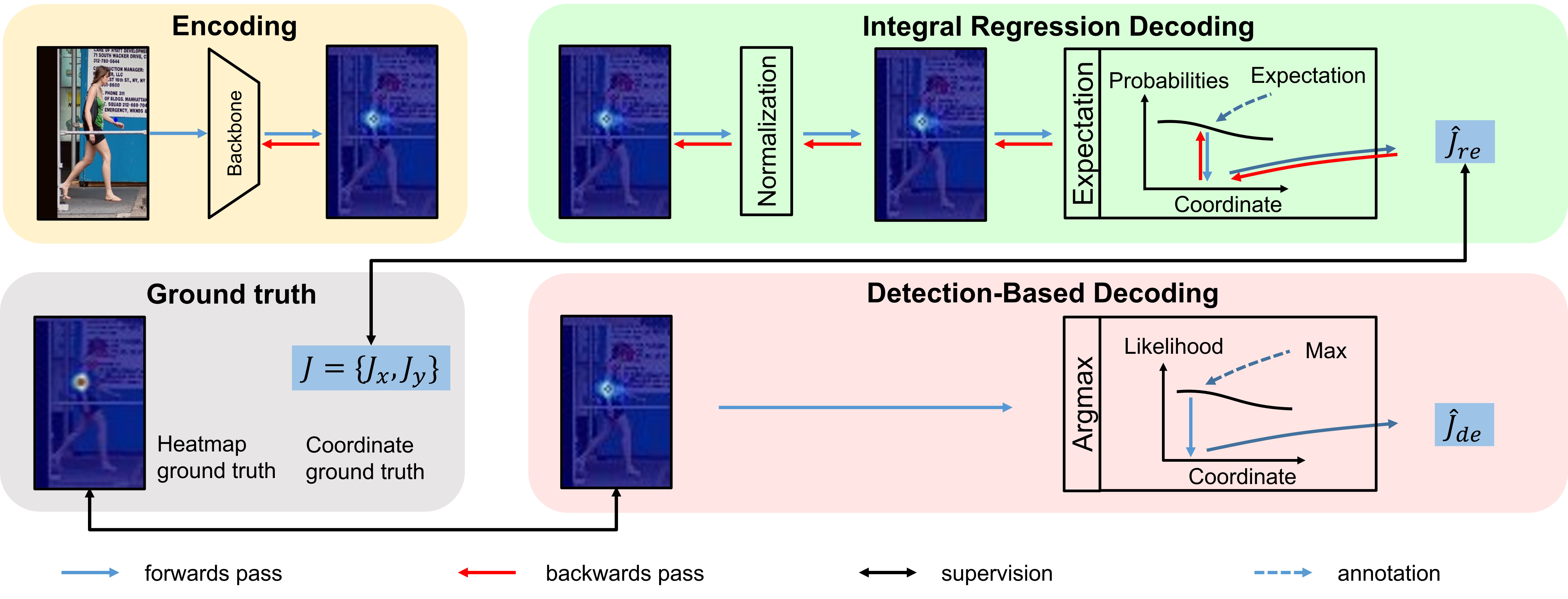}
    \caption{\small    
    Comparison of the two decoding processes for detection-based methods (pink area) and IPR methods (green area). They share the same encoding (yellow area) and ground truth (gray area) but have different decoding (green and pink areas). 
    }
    \label{fig:overview}
\end{figure*}

This work targets \emph{`top-down'} human pose estimation in which a person detector provides a bounding box around the person of interest.  We denote the cropped image of a given person as $I$ and focus our discussion on one of the $K$ joints in the body.  The pose estimation model outputs a heatmap $\hbH \in \mathbb{R}^{M\times N}$ where $M$ and $N$ are the dimensions of the spatial heatmap.  Typically, $M$ and $N$ are scaled down by a factor of $4$ from the input dimensions of $I$~\cite{xiao2018simple, sun2019deep}.  The heatmap $\hbH$ represents a discrete spatial likelihood $P(\bJ | I)$, where $\bJ \in \mathbb {R}^{1 \times 2}$ are 2D joint coordinates, \ie,  
\begin{equation}\label{eq:Jdef}
    \hbJ = \text{De}\left( P(\bJ | I ) \right) \approx \text{De}( \hbH ),
\end{equation}
\noindent and $\text{De}( \cdot )$ indicates the decoding operation. Both detection and IPR methods decode the coordinates $\hbJ$ from $\hbH$ (see Sec.~\ref{sec:img2heatmap}) but the two methods differ in their manner of decoding (see Sec.~\ref{sec:decode}) and the form of supervision for learning (see Sec.~\ref{sec:supervision}).

\subsection{Image to Heatmap Architectures}
\label{sec:img2heatmap}
Note that Eq.~\ref{eq:Jdef} is defined for a single joint for clarity. In practice, all $K$ heatmaps are predicted simultaneously by the same network, which we call the \emph{backbone}.  Each joint is one channel in a $K\times M \times N$ output from the backbone. 

Backbones that predict heatmaps are composed of two parts: one part decreases the resolution and a second part extracts feature representations at this reduced resolution~\cite{sun2019deep}.  Jointly, these two parts determine the quality of the visual representations, \ie, the heatmaps. For example, the HRNet backbone~\cite{sun2019deep} contains multiscale feature fusion subnetworks to get richer high-resolution information over the SBL series~\cite{xiao2018simple}, thus obtaining better results. For a fair comparison of the difference between detection and integral pose regression (IPR), the image to heatmap architectures or backbone should be the same. Unlike~\cite{sun2018integral}, which applied different backbones, our paper uses the same backbone.

\subsection{Heatmap Decoding: Max vs Expected Value}\label{sec:decode}

\textbf{Detection} methods apply an \emph{argmax} on $\hbH$ as the decoding operation $\text{De}(\cdot)$.  If the estimated heatmap $\hbH$ is indexed by $\bp$, the joint coordinates are estimated as $\hbJ_{\text{de}}$, where
\begin{equation}\label{eq:argmax}
    \hat{{\bJ}}_{\text{de}} = \underset{\bp}{\operatorname{argmax}} \; \hbH (\bp).
\end{equation}
\noindent Taking an {argmax} can be interpreted as taking a maximum likelihood on the heatmap $\hbH$, assuming that $\hbH$ is proportional to the likelihood.  In practice, the final $\hbJ_{\text{de}}$ value is determined as a linear combination of the highest and the second-highest response on $\hbH$.  This heuristic is meant to account for quantization effects in the heatmap as it is discrete, and it has shown to work better than the simple argmax~\cite{newell2016stacked}. The more recent work DARK~\cite{zhang2020distribution} approximated the true prediction with a Taylor series evaluated at the maximum activation of the heatmap, which further improved decoding accuracy. As the above heuristic shift and DARK are extensions of the basic argmax and remain non-differentiable, we consider the core argmax form for analysis in this paper.
\\

\noindent \textbf{Integral regression} uses an expectation operation as $\text{De}( \cdot )$ to estimate the joint coordinates. To take the expectation on the heatmap $\hbH$, $\hbH$ must first be normalized to sum up to 1. The most common (and effective~\cite{nibali2018numerical}) approach is to apply a softmax normalization. Afterward, the predicted joint coordinates $\hbJ_{\text{re}}$ with $x$ and $y$ components, $\hat{J}_x$ and $\hat{J}_y$\footnote{For clarity, we drop the subscript `$\text{re}$' when referring to the $x$ and $y$ components, 
as we refer to these individual components for integral regression only.}, are determined by taking the expectation on the normalized heatmap $\tbH$ with elements $\tilde{h}_{\bp}$ at location $\bp$:
\begin{equation}\label{eq:taking_expectation}
\hbJ_{\text{re}} =
    \begin{bmatrix}
        \hat{J}_x \\
        \hat{J}_y
    \end{bmatrix} = 
    \sum\limits_{\bp\in \Omega} \bp \cdot \tilde{h}_{\bp} ,
    \,\,\,\, \text{where} \,\,\, \tilde{h}_{\bp} = \frac{e^{\beta \hat{h}_{\bp}}}{\sum\limits_{\bp'\in \Omega} e^{\beta \hat{h}_{\bp'}}}.
\end{equation}

Here, $\Omega$ is the domain of the heatmap
and $\beta$ is a scaling parameter used in the softmax normalization.

\subsection{Supervision: Explicit Heatmap vs. Coordinates}\label{sec:supervision}
\textbf{Detection} methods are trained by providing explicit supervision on the heatmap. From a ground truth joint coordinate $\bJ$, a ground truth heat map $\bH$ can be generated by placing a circular Gaussian with the mean centered on $\bJ$ (see Fig.~\ref{fig:overview}) and a small standard deviation of a few pixels. 
For training, the loss applied is a pixel-wise MSE between the ground truth $h_{\bp}$ and the predicted $\hat{h}_{\bp}$:
\begin{equation} \label{eq:de_loss}
    L_{\text{de}} =  ||\bH - \hbH||_{2}^2 =  \sum_{\bp\in \Omega}(h_{\bp} - \hat{h}_{\bp})^2, 
\end{equation}
\noindent
where $\Omega$ is the domain of the heatmap.  Note that the loss in Eq.~\ref{eq:de_loss} is defined in terms of the heatmap and not the actual variable of interest, \ie, the predicted joints.  As such, detection methods are not end-to-end in their learning, and this is often cited as a drawback~\cite{sun2018integral, zhang2020distribution}.  \\

\noindent \textbf{Integral regression} defines a loss based on the difference between the prediction $\hbJ_{\text{re}}$ and ground truth joint location $\bJ$. While L2 and L1 losses are both applicable, L1 loss has empirically been shown to perform better than L2~\cite{sun2018integral}:
\begin{equation}\label{eq:re_loss}
    L_{\text{re}} = \| \bJ_{\text{gt}} - \hbJ_{\text{re}} \|_1 =(|\hat{J}_x - J_x | + |\hat{J}_y - J_y|).
\end{equation}

\noindent Because the loss in Eq.~\ref{eq:re_loss} is directly defined in terms of the joint coordinates, integral regression is considered to be an end-to-end method. As such, the estimated $\hbH$ is learned only implicitly, hence some works referred to the heatmap as ``latent''~\cite{iqbal2018hand}.

\section{A Localized Heatmap Model}

\begin{figure}[t]
\begin{center}
    \includegraphics[width=0.9\hsize]{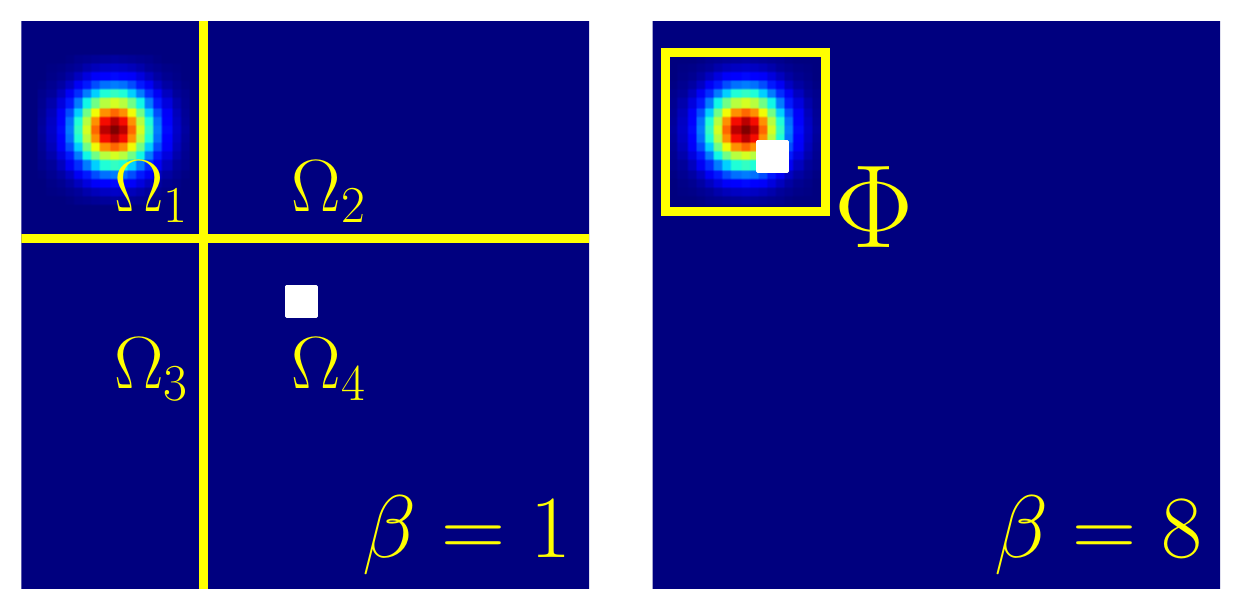}
\end{center}
\caption{Illustration of the integral regression bias and $\Phi$. Each image depicts a different implicit heatmap overlaid with the predicted joint location indicated by the white square. The bias, \ie, the difference between the heatmap mode and the predicted location, increases with smaller $\beta$. 
We propose a method to compensate for the bias by partitioning the heatmap into individual regions ($\Omega_1$ to $\Omega_4$, see left heatmap) to ensure that the heatmap mode is locally centered. Also, we assume the heatmap is only activated within $\Phi$ (see right heatmap).}
\label{fig:dists_betas}\label{fig:phi}
\end{figure}

First, we present a localized heatmap model (Sec.~\ref{sec:heatmap_model}), which serves as the foundation of our theoretical analysis.  When comparing the expected performance between the two methods (Sec.~\ref{sec:epe}), we found that integral regression should be at least comparable to if not better than detection, yet this is not observed in practice. We attribute this discrepancy to the localization collapse of the heatmaps in integral regression (Sec.~\ref{sec:localization_collapse}) and propose a simple regularizer to alleviate this problem (Sec.~\ref{sec:regularizer}). 

\subsection{Model Definition}\label{sec:heatmap_model}
To help with the theoretical analysis and comparison of the two methods, we make the following assumptions on the heatmap. 
First, we assume that a well-trained network produces a heatmap $\hat{h}_{\bp}$ with significant or large values in a \emph{localized} support region around $\hbJ$.  We denote the region of support as $\Phi$ and assume that outside of $\Phi$, the heatmap activation is approximately zero (see Fig.~\ref{fig:phi}).

Consider the normalized heatmap\footnote{For discussion purposes, we consider normalizing the heatmap for detection methods as per Eq.~\ref{eq:taking_expectation}. Note that this will not affect the outcome since $\underset{\bp}{\operatorname{argmax}}(\tbH)\!=\!\underset{\bp}{\operatorname{argmax}}(\hbH)$. } $\tbH$, which we assume can be modeled by some density distribution $\mathcal{P}$.  Note that we make no assumption on the form of $\mathcal{P}$, but we do assume that it fully captures the support for the joint location within $\Phi$. The $\Phi$ can be different among samples, \eg, easy samples are likely to have small $\Phi$s. Furthermore, we assume that the region of support $\Phi$ is centered on the expected value of $\mathcal{P}$, at location ${\bmu}=(\mu_x, \mu_y)$ with an area of $(2s+1, 2s+1)$, \ie,
\begin{multline}
    \bmu = \sum\limits_{\bp\in \Omega} \bp \cdot  \tilde{h}_{\bp} =  \sum\limits_{\bp\in \Phi(\bmu, s)} \bp \cdot  \tilde{h}_{\bp} \\ =\begin{bmatrix}
    \sum\limits_{i=\mu_x-s}^{\mu_x+s}\sum\limits_{j=\mu_y-s}^{\mu_y+s} i \cdot \tilde{h}_{ij},
    \sum\limits_{i=\mu_x-s}^{\mu_x+s}\sum\limits_{j=\mu_y-s}^{\mu_y+s} j \cdot \tilde{h}_{ij}
\end{bmatrix}^{\intercal}. 
\end{multline}

We verify this localized heatmap model (see detailed experiments in Sec.~\ref{sec:exp_heatmapmodel}) by tallying the activations within the support region $\Phi$. In a heatmap of $64\times48$, a $19\times 19$ region contains 95\% of all non-zero activations for both detection and integral regression. 

\subsection{Expected End-Point Error (EPE)}
\label{sec:epe}

Based on the definitions in the localized heatmap, we can estimate an expected end-point error (EPE). For detection methods, which decode the heatmap with an argmax, we assume that each coordinate position ${\bp}$ has some probability $w(\bmu, s)$ of being the maximum activation, \ie, argmax$(\mathcal{P})_{\Phi(\bmu, s)}\!\sim\! w(\bmu, s)$. Here we make an additional assumption that $w(\bmu, s)$ represents a radially symmetric distribution centered on $\bmu (\mu_x, \mu_y)$ with all non-zero support contained within $\Phi$. It follows that the expected EPE for a joint
can be defined as:
\begin{equation}\label{eq:epe_det}
    \mathbb{E}_{\text{de}} (\bmu, s)  = \!\!\! 
    \sum_{\bp \in \Phi(\bmu, s)}\!\!\! w (\bp) ||\bJ_{\text{gt}} -\bp ||_2, \,\,\,\sum w(\bp) =1.
\end{equation}
For integral regression methods, which decode the heatmap with an expectation, the estimated joint coordinate aligns with the center of $\Phi$ by definition, \ie, $\hbJ\!=\!\bmu$, leading to the following expected EPE for a single joint: 
\begin{equation}\label{eq:epe_reg}
\mathbb{E}_{\text{re}} (\bmu) = ||\bJ_{\text{gt}} -\bmu ||_2=\!\!\!\sum_{\bp \in \Phi(\bmu, s)}\!\!\!w(\bp)||\bJ_{\text{gt}} -\bmu ||_2, \sum w(\bp) =1.
\end{equation}

\noindent
In both Eq.~\ref{eq:epe_det} and Eq.~\ref{eq:epe_reg}, the heatmap density $\Phi(\bmu, s) =\{\bp(\mu_x+i, \mu_y+j)\,\,|\,\,\ \forall i,j \in [-s, s]\}$. Due to the centrosymmetry of $\Phi(\bmu, s)$, $\Phi(\bmu, s)$ can be indexed by 
$\bmu$ and pairs of points ($\bp_1, \bp_2$) where $\bp_1+\bp_2=2\bmu$. Additionally, $\forall(\bp_1, \bp_2 \in \Phi(\bmu, s)) \wedge (\bp_1+\bp_2=2\bmu)$, we have the following:
\begin{multline}\label{eq:triangle_inequality}
    \!\!\!\!w(\bp_1)||\bJ_{\text{gt}}-\bp_1|| + w(\bp_2)||\bJ_{\text{gt}}-\bp_2|| \geqslant  w(\bp_1) ||2\bJ_{\text{gt}}-\bp_1-\bp_2||\\= (w(\bp_1)+w(\bp_2))||\bJ_{\text{gt}}-\bmu||,
\end{multline}
\noindent
where $w(\bp_1) = w(\bp_2)$. Summing up all the centrosymmetric $\bp$ pairs and $\bmu$ on both sides of Eq.~\ref{eq:triangle_inequality}, we see that Eq.~\ref{eq:epe_det} is the summation of left hand side of Eq.~\ref{eq:triangle_inequality} while Eq.~\ref{eq:epe_reg} is the summation of right hand side. 
Therefore, when $s>0$, for a fixed  $(\bmu, s)$, the expected EPE of detection is greater than that of regression, leading to worse accuracy.

Experimentally, we verified that in hard cases, where both detection and integral regression present a large $s$ around ground truth location, the expected EPE of detection is greater than that of integral regression. Details of the experiments are provided in Sec.~\ref{sec:comp_results}.

\subsection{Density Shrinkage \& Collapse}
\label{sec:localization_collapse}

The results of Sec.~\ref{sec:epe} suggest that for some fixed $(\bmu, s)$, the expected EPE of regression should always be lower than the expected EPE of detection. Yet in our findings, integral regression performs better only on the ``hard'' cases. This implies that the two methods do not predict heatmaps with the same $\bmu$ and $s$ on all the samples. This raises the question of how different heatmap distributions influence the predicted pose accuracy. We further investigate by comparing the predicted heatmap distributions and ground truth annotations.

As reported in~\cite{cocowebsite}, there exists a standard deviation for each keypoint among annotations by different human annotators and the human-annotated keypoints are normally distributed.
In this way, we can assume a true location of the keypoint $\bmu_{\text{true}}$ and that the annotated ground truth coordinate $\bJ_{\text{gt}}$ follows a Gaussian distribution centered at $\bmu_{\text{true}}$ with standard deviation $\sigma$:
\begin{equation}
    \bJ_{\text{gt}} \sim \mathcal{N}(\bmu_{\text{true}}, \sigma_{\text{true}}).
\end{equation}

Based on our proposed localized heatmap model, we further assume that the radial distribution of prediction in Sec.~\ref{sec:epe} is Gaussian. 
Specifically, we assume that the spatial probability of the joint location $\hbJ$ follows a Gaussian distribution centered at $\bmu$ with standard deviation $\hat{\sigma}$:
\begin{equation}
    \hat{\bJ} \sim \mathcal{N}(\bmu, \hat{\sigma}) = \mathcal{N}(\bmu_{\text{true}}+(\bmu-\bmu_{\text{true}}), \hat{\sigma}),
\end{equation}

As the model is trained by the ground truth label $\bJ_{\text{gt}}$, the prediction error is minimized when the distribution of predicted joints is similar to or matches the ground truth distribution. Applying the commonly used metric \emph{Bhattacharyya distance} $D_B$ to measure the distribution similarity of $\bJ_{\text{gt}}$ and $\hbJ$, we arrive at
\begin{equation}\label{eq:db}
    D_B(\bJ_{\text{gt}}, \hbJ, \hat{\sigma}^2) = \frac{1}{4}\text{ln}(\frac{1}{4}(\frac{\sigma_{\text{true}}^2}{\hat{\sigma}^2}+\frac{\hat{\sigma}^2}{\sigma_{\text{true}}^2}+2))+\frac{1}{4}(\frac{(\bmu-\bmu_{\text{true}})^2}{\sigma_{\text{true}}^2+\hat{\sigma}^2}).
\end{equation}

\noindent The optimal $\hat{\sigma}^*$ is achieved by minimizing the $D_B$:
\begin{equation}
    \sigma^* = \underset{\hat{\sigma}}{\operatorname{argmin}}(D_B(\bJ_{\text{gt}}, \hbJ, \hat{\sigma})).
\end{equation}

\noindent Setting the first derivative to 0 (see details in Appendix~A), $\hat{\sigma}^*$ will be constrained by the following:
\begin{equation}\label{eq:optimal_sigma}
    \frac{4}{(\frac{\sigma_{\text{true}}^2}{{\hat{\sigma}^*}{}^2}+\frac{{\hat{\sigma}^*}{}^2}{\sigma_{\text{true}}^2}+2)}(\frac{1}{\sigma_{\text{true}}^2}-\frac{\sigma_{\text{true}}^2}{{\hat{\sigma}^*}{}^4})=(\frac{\bmu-\bmu_{\text{true}}}{\sigma_{\text{true}}^2+{\hat{\sigma}^*}{}^2})^2.
\end{equation}

\noindent In the special case where $\hbJ\! =\!\bmu_{\text{true}}$, the optimal $\hat{\sigma}^* = \sigma_{\text{true}}$ will minimize the distance. However, under most conditions, $\bmu$ and $\bmu_{\text{true}}$ are not the same.  For 
Eq.~\ref{eq:optimal_sigma} to hold, the optimal $\hat{\sigma}^*$ must be greater than $\sigma_{\text{true}}$.
Since the right hand side of Eq.~\ref{eq:optimal_sigma} is always positive, one can derive from the left hand side that the optimal $\hat{\sigma}^*$ must be greater than $\sigma_{\text{true}}$.

Furthermore, when $0\!<\!\hat{\sigma}\!<\!\hat{\sigma}^*$, then $D_B^{'}(\hat{\sigma})<0$.  On the other hand, when $\hat{\sigma}\!>\!\hat{\sigma}^*$, then $D_B^{'}(\hat{\sigma})\!>\!0$. 
Under this circumstance, an extremely localized heatmap with a very small $\hat{\sigma}$ is likely to have worse performance compared with less localized heatmaps with a slightly larger $\hat{\sigma}$.

We empirically verified the above theory on both detection and integral regression. For detection, we supervised predicted heatmaps with Gaussian heatmaps centered at the ground truth location with different standard derivations. Results revealed that Gaussian heatmaps with $\sigma\!=\!2$ best match the distribution of ground truth coordinate $\bmu_{\text{true}}$. A smaller or a larger $\sigma$ worsens the performance. For integral regression, the overall standard deviation is smaller than that of detection ($\sigma\!=\!2$). When applying a heatmap shrinkage regularizer (in Sec.~\ref{sec:regularizer}) to enlarge the standard deviation, the performance is improved. The experiments are detailed in Sec.~\ref{sec:exp_extremelocalize}.

We further investigated the influence of heatmap diversity on prediction accuracy by separately evaluating the performance on easy/medium/hard cases. For both methods, there exists a density shrinkage such that when the samples transition from ``hard'' to ``easy'', $s$ and $\sigma$ get smaller. Especially for ``easy'' cases, $\sigma$ of detection follows the ground truth value (\eg, $\sigma\!=\!2$) while integral regression has an extremely small $\sigma$. In this case, $\sigma$ of integral regression is much smaller than the optimal $\hat{\sigma}^*$, and the performance is worse than detection. However, in ``hard'' cases where detection and integral regression present similar $\sigma$, the performance of integral regression is better than detection, as analyzed in Sec.~\ref{sec:epe}. Experimental details are provided in Sec.~\ref{sec:exp_localcomp}.

\subsection{Heatmap Shrinkage Regularization}
\label{sec:regularizer}
To prevent predicted heatmaps from being extremely localized, the value of adjacent pixels should not differ too much. Therefore, we propose a simple regularizer that penalizes large values of the heatmap processed by a $3\times3$ Laplacian filter. A Laplacian filter is preferred as it involves the center pixel along with its four adjacent pixels.
Specifically, the regularizer is given as follows:
\begin{equation}\label{eq:regularizer}
    L_{reg} = \sum_{\tilde{h}\in \tbH}(|\nabla^2 \tilde{h} - \tau | + \nabla^2 \tilde{h} - \tau).
\end{equation}
\noindent
In Eq.~\ref{eq:regularizer}, $|\cdot|$ is the absolute value function and $\tau$ is a hyperparameter controlling the spatial extent of the normalized heatmap. The regularizer loss is summed over all pixels convolved with the Laplacian filter. Note that the convolved value will not be penalized when below the controlling hyperparameter $\tau$.

\section{The Softmax Bias of Integral Regression}

In investigating why integral regression presents extremely localized heatmaps, we discovered that there is an induced bias when combining softmax and expectation operations (see Sec.~\ref{sec:bias}). Due to the presence of this bias, heatmaps with large standard deviations will shift the prediction from the localized center $\bmu$ toward the center of the image. Only when heatmaps are extremely concentrated around the ground truth location will the expected value be the ground truth coordinate. To alleviate the extremely localized heatmap and obtain the correct coordinate, we propose a simple scheme to compensate for the bias (see Sec.~\ref{sec:bias_compensation}). In Sec.~\ref{sec:proposed_method}, we address both  limitations mentioned in Sec.~\ref{sec:localization_collapse} and Sec.~\ref{sec:bias} and propose the Bias Compensated Integral Regression (BCIR), a regression-based framework that incorporates Gaussian prior loss to speed up training and improve prediction accuracy.

\subsection{Bias Definition}\label{sec:bias}

To take the expectation, we need a normalized probability density function; the softmax serves that purpose to normalize ${\hat{\bH}}$.  However, the softmax is also dense in that it assigns nonzero values to \emph{all} the pixels in $\tilde{\bH}$, even for the zero elements of ${\hat{\bH}}$\footnote{Consider the numerator of Eq.~\ref{eq:taking_expectation}, $\exp(\beta \cdot 0) = 1$}.  The non-zero assignments to the (close to) zero-valued pixels of $\hat{\bH}$ in turn contribute to the expected value and bias the estimated coordinate $\hat{\bJ}_{\text{re}}$ toward the center of the heatmap.  The further away the joint coordinate is from the center, the greater the bias (see Fig.~\ref{fig:dists_betas}).

The effects of such a bias can be alleviated somewhat by choosing an appropriate value for $\beta$ in the softmax. The smaller the $\beta$, the more the function distributes the probability density in the heatmap, hence greater impact of the zero-pixels in $\hat{\bH}$.  The larger the $\beta$, the more the function concentrates the density around the largest values of $\hat{\bH}$.  In the limit when $\beta$ goes to infinity, the softmax converges to the argmax function~\cite{chapelle2010gradient} and becomes non-differentiable.  As $\beta$ gets progressively larger, the gradients of the pixels that are further away from the center become smaller and gradually approach zero.  It is therefore necessary to make a tradeoff between the extent of the bias and having sufficient gradients for learning. Note that learning an extremely localized heatmap is equivalent to having a large $\beta$ in the forward propagation. However, as introduced in Sec.~\ref{sec:localization_collapse}, it harms the performance of the integral regression method.

\subsection{Bias Compensation}
\label{sec:bias_compensation}
As shown in Fig.~\ref{fig:dists_betas}, there is no bias only when the probability density is centered on the heatmap.  A naive way to compensate for the bias is to shift the coordinate system and center the heatmap at the ground truth coordinate $\bJ_{\text{gt}}$; then there would be no bias when taking the expectation of $\tilde{\bH}$. However, this requires knowing the location of $\bJ_{\text{gt}}$, which is feasible for training, but not inference.

As such, we propose a bias compensation scheme that removes the contributions of the additional support encoded in $\hat{\bH}$, \ie, extra non-zero assignments.  Suppose we wish to recover the true coordinate location $(x_o, y_o)$,  and assume for now that $(x_o, y_o)$ lies in the upper left quadrant of the heatmap, we can partition the image plane into four rectangular sections with splits at $2x_o$ and $2y_o$, with $\Omega_1$ as the section that contains $(x_o, y_o)$ and $\Omega_2$, $\Omega_3$, and $\Omega_4$ denoting the other regions in a clockwise fashion (see the left image in Fig.~\ref{fig:dists_betas}).  Based on this partition, we can split the expectation defined in Eq.~\ref{eq:taking_expectation} as follows:
\begin{equation}~\label{eq:splitexpectation}
    \hat{\bJ}_{\text{re}} = \sum\limits_{\mathbf{p} \in \Omega_1}\Tilde{\mathbf{H}}(\mathbf{p})\cdot\mathbf{p} \,\,\,\, +  \!\!
    \sum\limits_{\mathbf{p} \in (\Omega_2,\Omega_3,\Omega_4)} \!\!\!\! \Tilde{\mathbf{H}}(\mathbf{p})\cdot\mathbf{p}.
\end{equation}
We assume that the support for $(x_o,y_o)$ is well localized, \ie, fully contained within $\Omega_1$ in $\bH$ and that sections $\Omega_2$ to $\Omega_4$ contain only zero or near-zero elements.   As such, only the first term of Eq.~\ref{eq:splitexpectation} should contribute to  $\hat{\bJ}_{\text{ro}}$.  It follows then that the joint location can be estimated as a scaled version of the first term of Eq.~\ref{eq:splitexpectation} :
\begin{equation}
     \hat{\bJ}_{\text{ro}} = \! \frac{1}{w_1}\!\sum\limits_{\mathbf{p} \in \Omega_1}\Tilde{\mathbf{H}}(\mathbf{p})\cdot\mathbf{p}, \,\, \text{where } w_1\! =\!\sum\limits_{\mathbf{p} \in \Omega_1}\Tilde{\mathbf{H}}(\mathbf{p}),
\end{equation}
\noindent where we define $\hat{\bJ}_{\text{ro}}$ as the estimate of $(x_o,y_o)$, \ie, a  bias-compensated joint location. Note that the above formulation is implicit since $\Omega_1$ depends on $(x_o,y_o)$. With algebraic rearrangement, we can formulate $\hat{\bJ}_{\text{ro}}$ as a function of $\hat{\bJ}_{\text{re}}$:
\begin{align}
\hat{\bJ}_{\text{ro}} & = \tfrac{C}{(C-wh)} \hat{\bJ}_{\text{re}} -
    \begin{bmatrix}
    \frac{hw^2}{2(C-wh)} \\ 
    \frac{h^2w}{2(C-wh)}
    \end{bmatrix}.
    \label{eq:linear-bias}
\end{align}

\noindent Above, $C$ is the normalizing constant used in the softmax, \ie, the denominator of Eq.~\ref{eq:taking_expectation}, and is a function of $\beta$: 
\begin{equation}\label{eq:denomsoftmax}
     C(\beta) = \sum\limits_{\mathbf{p} \in \Omega}\exp(\beta \cdot \hat{\bH}(\mathbf{p})). 
\end{equation}
\noindent From Eq.~\ref{eq:linear-bias} and Eq.~\ref{eq:denomsoftmax}, we see that the impact of the bias is negligible for a large $C$ since the scaling factor approaches one while the offset approaches zero.  This is exactly the case when a large $\beta$ is used, \ie, the softmax approaches the argmax function.  However, when $C$ is small, then the bias becomes more significant; so if $\beta$ is not sufficiently large, the network must learn very large and concentrated values of $\hat{\bH}(\bp)$ to estimate the correct $\hat{\bJ}_{\text{re}}$. We posit that it is exactly this interplay that makes it very challenging for the network to learn, and hence the slow convergence rates of the integral regression method (see Sec.~\ref{sec:ablation_study}).

Eq.~\ref{eq:linear-bias} recovers the bias-compensated joint location $\hat{\bJ}_{\text{ro}}$ without any knowledge of the ground truth.
We refer the reader to Appendix~B for the full derivation and the cases when $(x_o, y_o)$ are in the other quadrants.  During inference, we can directly compensate for the biased location $\hat{\bJ}_{\text{re}}$ based on the expectation in Eq.~\ref{eq:taking_expectation}.  We can do the same during training and simply update the L1 loss of Eq.~\ref{eq:re_loss} with $\hat{\bJ}_{\text{ro}}$, \ie,
\begin{equation} \label{eq:unbiased_rergession}
    \mathcal{L}_{\text{re}} = \|\hat{\bJ}_{\text{ro}} - \bJ_{\text{gt}}\|_1.
\end{equation}
In our approach, we have opted to retain the softmax and instead compensate for the expectation. Naive activation functions in place of the softmax have been explored for human pose estimation but have been shown to be less effective~\cite{nibali2018numerical}. A less naive option is the sparsemax~\cite{martins2016softmax}, which has been proposed as a sparse alternative to the softmax.  It projects the pre-activation value to a simplex so only a few non-zero values are preserved. However, given the large size of the flattened heatmap, only assigning a number of pixels with non-zero values makes it hard to train.

\subsection{Bias Compensated Integral Regression (BCIR)}
\label{sec:proposed_method}

Facing the two limitations that integral regression exhibits -- an extremely localized heatmap and an induced bias -- we propose a new method to alleviate the limitations by regularizing the heatmap and incorporating a bias compensation.

The proposed regularizer in Sec.~\ref{sec:regularizer} helps flatten the heatmap.  However, it does not indicate the ground truth location of the heatmap. Inspired by the spatial supervision in the detection-based methods~\cite{xiao2018simple, sun2019deep}, we find that adding a spatial prior further boosts the performance. It not only prevents heatmap collapse but also provides denser supervision. The loss can be concluded as
\begin{equation}
    \mathcal{L} = \mathcal{L}_{\text{re}} + \lambda (t) \cdot 
    L_{\text{de}},
\end{equation}
\noindent where we use a simple step function for $\lambda(t)$, \ie, $\lambda(t)\!=\!1$ for $t<T_o$ and $\lambda(t)\!=\!0$ for $t>=T_o$. Compared to detection-based methods which supervise heatmaps with fixed Gaussian heatmaps, $\lambda(t)$ has the effect of minimizing the impact $L_{\text{de}}$ in subsequent epochs and allows the network to learn the arbitrary shape of the implicit heatmap in the final stage. The key purpose of our method is to let the network learn an unbiased heatmap and make the distribution of the predicted heatmap closer to the true distribution of the ground truth locations. During inference, the unbiased coordinate $\hat{\bJ}_{\text{ro}}$ is final output of our network.

\section{Gradients of Integral Regression}\label{sec:training_analysis}
The difference in heatmap decoding not only influences the final coordinates given the same assumed heatmap, but also determines the gradients that arise from the loss. The gradients in turn influence the learning and thereby the generation of the heatmaps themselves. 
In this section, we explore the gradients of heatmaps with respect to the loss function for detection and regression-based methods and pinpoint the specific gradient components that slow down the learning of integral regression.  

\subsection{Heatmap Gradients}

\noindent
\textbf{Detection.} The gradient of $L_{\text{de}}$ (see Eq.~\ref{eq:de_loss}) with respect to each pixel in the estimated heatmap $\hat{h}_{\bp}$ is straightforward:
\begin{equation} \label{eq:de_backward}
    \frac{\partial L_{\text{de}}}{\partial \hat{h}_{\bp}} = 2(\hat{h}_{\bp} - h_{\bp}).
\end{equation}
The gradient in Eq.~\ref{eq:de_backward} features the predicted heatmap value $\hat{h}_{\bp}$ and the ground truth heatmap value $h_{\bp}$ and provides explicit supervision at every pixel.  False positives \emph{and} false negatives are penalized by reducing wrong high likelihoods and raising incorrect low likelihoods, respectively. Specifically, for each position $\bp$ in the heatmap, if the predicted $\hat{h}_{\bp}$ is smaller than the ground truth $h_{\bp}$, the gradient is negative and proportionally increases the value in the next iteration, and vice versa for the predicted $\hat{h}_{\bp}$ larger than the ground truth.\\

\noindent
\textbf{Integral Regression.} The gradient of the loss $L_{\text{re}}$ (see Eq.~\ref{eq:re_loss}) with respect to $\hat{h}_{\bp}$ can be estimated based on the chain rule (see detailed derivation in Appendix~C) as
\begin{multline}\label{eq:chain_rule}
\!\!\nabla_{\bp}\! :=
\frac{\partial L_\text{re}}{\partial \hat{h}_{\bp}} = 
\frac{\partial L_\text{re}}{\partial \hbJ}
\frac{\partial \hbJ}{\partial \tilde{h}_{\bp}}
\frac{\partial \tilde{h}_{\bp}}{\partial \hat{{h_{\bp}}}}\\ = \,\,\,\,\beta\!\!\!\!\!\!\!\!\!\!\! \underbrace{ \tilde{h}_{\bp}}_{\nabla_1 \;\text{(value factor)}} \!\!\!\!\!\!\!\!\!\! \underbrace{ ( \text{s}(\hat{J}_x - J_x)(i-\hat{J}_x)+ \text{s}(\hat{J}_y - J_y)(j-\hat{J}_y))}_{\nabla_2 \;\text{(location factor)}}.
\end{multline}
In Eq.~\ref{eq:chain_rule}, the gradient can be split into two terms of interest: $\nabla_1$ and $\nabla_2$, which we name the value factor and the location factor, respectively.  The nature of the value and location factor gives strong indications to the learning process for integral regression methods.  Firstly, the gradient $\nabla_{\bp}$ is proportional to the normalized predicted heatmap value $\tilde{h}_{\bp}$ as given in the value factor. This factor makes it prone to gradient vanishing wherever locations ${\bp}$ have small heatmap predictions.  Secondly, the location factor is a linear combination of $i$ and $j$, which present a linear plane where faraway points like corners are more likely to have large magnitudes. Though these two points do not prevent the loss from decreasing during learning, they slow down training. We further elaborate this in Sec.~\ref{sec:reg_analysis}.

\subsection{Updating criterion of IPR}
\label{sec:reg_analysis}
We now conduct a detailed analysis of the gradients presented in Eq.~\ref{eq:chain_rule} and show the typical cases that arise during learning.   When referring to $\Omega$, the domain of the heatmap, we use a coordinate system with the origin at the upper left corner and increasing coordinates moving to the lower right corner (see Fig.~\ref{fig:experimental_support}(a)). For simplicity of discussion, we assume that the ground truth joint $\bJ_{\text{gt}}$ is located somewhere in the lower right quadrant of $\Omega$, though the analysis holds for $\bJ_{\text{gt}}$ in all other quadrants of $\Omega$ analogously. The visualizations of the {four scenarios} can be viewed in the left panel of Fig.~\ref{fig:rede_toy}.

Let us suppose $\bH_n=g(\bw_n)$, where $\bH_n$ is the heatmap of a given image $I$ at the $n$-th iteration, $g(\bw_n)$ is the function of generating the heatmap, \ie, the backbone network, with weights at the $n$-th iteration. We approximate the updating of a heatmap by

\begin{equation}\label{eq:h_update}
h_{\bp}^{n+1} \approx h_{\bp}^{n} - \gamma\nabla_{\bp}^n,
\end{equation}

\noindent where $\gamma$ is a constant, $h_{\bp}^n$ and $\nabla_{\bp}^n$ denote the heatmap and the gradient value located at $\bp$ in the heatmap at the $n$th iteration. $h_{\bp}^0$ denotes the original heatmap. The detailed derivation in given Appendix~D. A larger magnitude of $\nabla_{\bp}$ has a greater change at the location $\bp$ in the next iteration.
As the value and location factors have different contributions to the gradient from Eq.~\ref{eq:chain_rule}, and these are in turn affected by the size $s$ and location $\mu$ of $\Phi$, we can define four characteristic cases of $\Phi$ and outline how learning is affected in each case.

\noindent \textbf{(1) $s$ is large, $\Phi$ has uniformly random values}, \eg, a randomly initialized heatmap at the start of training.  As $\tilde{h}_{\bp}$ is in a similar range for all ${\bp}$, the value factor $\nabla_1$ (see Eq.~\ref{eq:chain_rule}) is  approximately similar for all pixels. As such, the distinction between the gradient of one pixel and that of another is determined by the location factor $\nabla_2$.  Given that the ground truth coordinates $(J_x, J_y)$ are in the lower right quadrant, as long as $(J_x, J_y)$ are both greater than the predicted coordinates $(\hat{J}_x, \hat{J}_y)$, $\nabla_{\bp}$ has a {gradation} that increases progressively toward the bottom right corner (Fig.~\ref{fig:rede_toy} Regression Case 1). Generally, the location factor pushes the activations of the heatmap toward the corner of the correct quadrant as corners own the largest gradients in the resultant linear plane of the location factor $\nabla_2$.\\

\noindent \textbf{(2) $s$ is small, $\mu$ is far from the ground truth $\bJ$}, \eg, when $\mu$ is in the upper left quadrant of $\Omega$ while the ground truth $\bJ_{\text{gt}}$ is at the bottom right. A typical scenario is when there are activations in the heatmap around the left ankle for the ground truth right ankle. In this case, for the pixels outside of $\Phi$, the value factor approaches zero, \ie, $\tilde{h}_{\bp} \xrightarrow{} 0$, and pushes $\nabla_{\bp}$ toward zero so the pixels receive very limited gradient updates. For the pixels in $\Phi$, the values decrease gradually until the value factor outside of $\Phi$ no longer dominates, \ie, they reach the same scale.  At this point the heatmap will return to case (1). \\

\noindent \textbf{(3) $s$ is small, $\mu$ is in the corner of the same quadrant as the ground truth $\bJ_{\text{gt}}$.}
Initially, this case is similar to case (2). In the process of all the elements becoming the same scale, the activations move diagonally (in this case toward the upper left quadrant), and the prediction approaches the ground truth coordinate (See Fig.~\ref{fig:rede_toy} Case 3).\\

\noindent \textbf{(4) $s$ is small, $\mu$ is close to the ground truth $\bJ_{\text{gt}}$.}  
This scenario occurs when the model is reasonably trained and can roughly localize the joint. The gradient at the ground truth pixel $\bJ_{\text{gt}}$ is always negative or zero, \ie,
{
\begin{multline}\label{eq:gradients_gt}
    \nabla_{(J_x, J_y)} =-\beta  \tilde{h}_{(J_x,J_y)}(\|\hat{J}_x - J_x\|+ \|\hat{J}_y - J_y\|).
\end{multline}
}

\noindent This non-positive value guides the network to predict a large heatmap response at the ground truth location, \ie, a large $ \tilde{h}_{(J_x,J_y)}$. {With the bias introduced in Sec.~\ref{sec:bias}, the predicted joint locations $\hat{J_x}$ and $\hat{J_y}$ will not overlap with the ground truth $J_x$ and $J_y$.  This in turn pushes the network to predict a few extraordinarily large activations around the ground truth location. As such, the support region $\Phi$ shrinks and leads to a very small $\sigma$, as shown in Sec.~\ref{sec:localization_collapse}. We verify this by compensating the bias (Eq.~\ref{eq:unbiased_rergession}) in Table~\ref{tab:ablation} (line +de-bias), which shows that once the bias compensation is added, the heatmaps are less localized.}

\section{Experiments}

\subsection{Datasets and Implementation}
\label{sec:dataset}

\subsubsection{Datasets and Evaluation Metrics}
 We evaluate on two human pose datasets, MS COCO~\cite{lin2014microsoft} and MPII~\cite{andriluka20142d}, and one hand pose dataset, RHD~\cite{zimmermann2017learning}. 

The \textbf{COCO dataset} has 250k person instances with 17 annotated keypoints. We evaluate with the standard metric, Object Keypoint Similarity (OKS).  OKS normalizes the absolute error between the predicted location and the ground truth location with the size of the person.  We use the primary challenge evaluation metric, mean average precision (AP), over 10 OKS thresholds to evaluate the performance. We also report the value before normalization, the squared Euclidean distance between the prediction and ground truth, which we denote as End-Point Error (EPE).

The \textbf{MPII dataset} contains 49k person instances with 16 annotated keypoints. We use the standard train/validation split of~\cite{tompson2014joint} and evaluate performance with Percentage of Correct Keypoints (PCK) and EPE.  

\textbf{RHD} is a synthetic hand dataset with 41k training and 2.7k testing images from 20 animated characters. For each RGB image, 21 hand keypoint annotations are provided.  We follow~\cite{iqbal2018hand} and evaluate with AUC and EPE.\\

\subsubsection{Implementation Details}
We implement our experiments in Pytorch and train the networks with the Adam optimizer. 
As baseline models, we use SBL~\cite{xiao2018simple} and HRNet~\cite{sun2019deep} with different backbones, \eg, SBL-ResNet50 and HRNet-W32, and follow the same learning configurations for detection-based, regression-based, and BCIR. For a fair comparison, we rerun the experiments, \ie, detection-based methods, to report the results. {We use the same image to heatmap architecture with a fixed input size to generate the heatmap $\mathbf{H}$ for all $K$ joints. FasterRCNN~\cite{ren2015faster} is applied to detect all human instances, and all bounding boxes cropped from the original images with different resolutions are resized to the input size of the network.} From Eq.~\ref{eq:taking_expectation}, we set $\beta\!=\!10$ as a default value. 
For the proposed BCIR, we use a $T_o$ of 120 epochs for SBL and 190 epochs for HRNet. {Unless specified otherwise, the experimental verification in Secs.~\ref{sec:exp_heatmapmodel}-\ref{sec:gradexp}
and Sec.~\ref{sec:ablation_study} are implemented with the SBL-ResNet50 backbone, trained on the MSCOCO training set and tested on the MSCOCO validation set. The input size is $256 \times 192$.}

\subsection{Sub-Benchmark Performance Comparison}
\begin{figure}[t]
    \centering
    \includegraphics[width=0.5\textwidth]{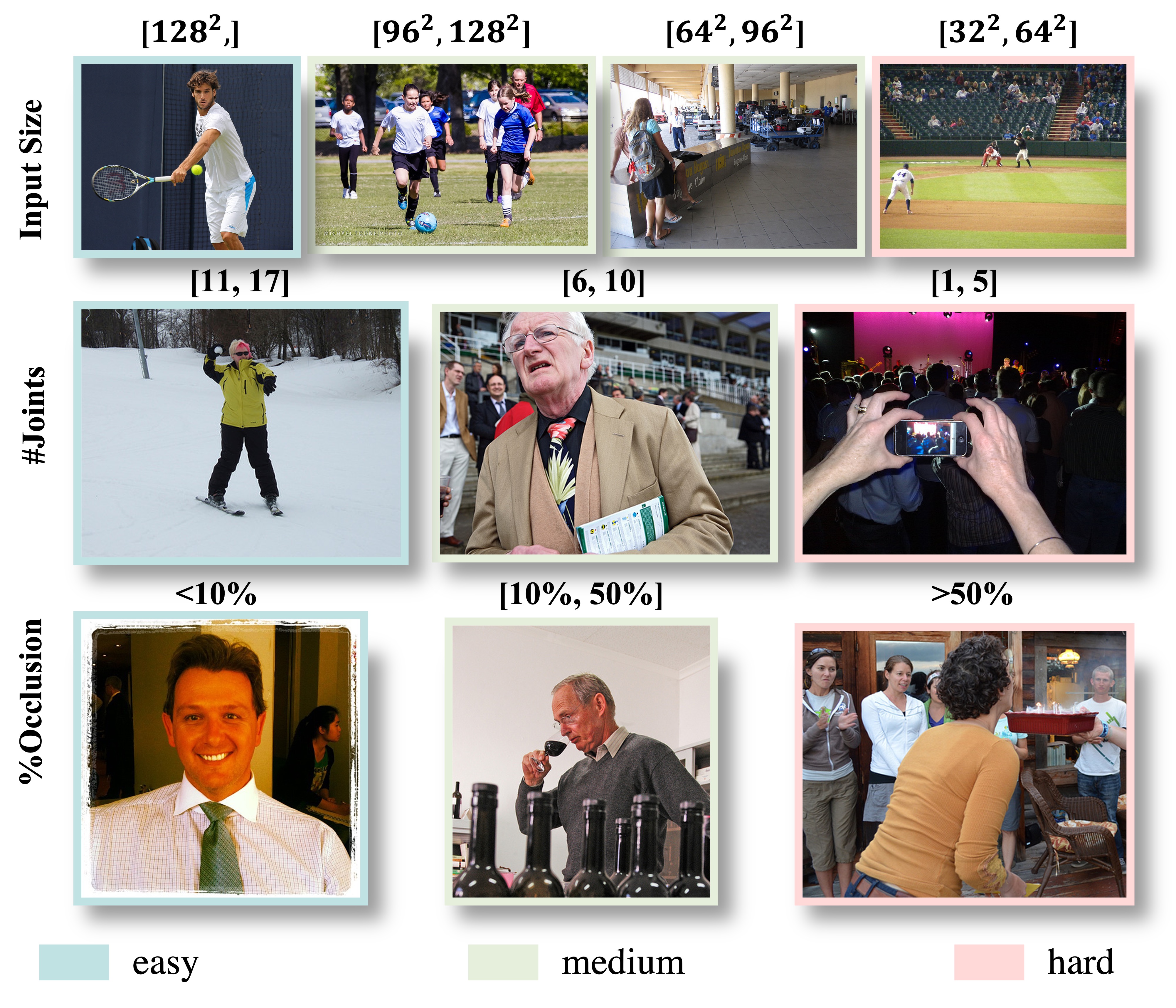}
    \caption{Samples from the COCO dataset separated by their input (bounding box) size, number of joints present in the scene, and the percentage of occlusion (of the present joints). 
    The color of the image border indicates the sample difficulty.
    }
    \label{benchmarking_samples}
\end{figure}

\subsubsection{Dataset Separation}\label{factors}
We divide the COCO val set according to the three factors outlined in~\cite{ruggero2017benchmarking}: the number of joints or keypoints present in the scene (11-17, 6-10, 1-5), the percentage of occlusion ($<\!\!10\%$, 10-50\%, $>\!50\%$), and the largest dimensions of the input bounding box ($>\!\!128$px, 96-128px, 64-96px, 32-64px). Based on this division, we further label the cases as \emph{``easy''} (11-17 joints, $<\!\!10\%$ occlusion, $>\!\!128$ px input), \emph{``medium''} (6-10 joints, 10-50$\%$ occlusion, 64-128px input), and \emph{``hard''} (1-5 joints, $>\!50\%$ occlusion, 32-64px input). Fig.~\ref{benchmarking_samples} shows some visual examples of these cases.

\begin{table}[t]
\footnotesize
    \centering
    \scalebox{0.85}{
    \begin{tabular}{l|c c c|c}
    \hline
    & \multicolumn{3}{c}{\# Joints} & \multicolumn{1}{|c}{}\\ 
    \hline\hline
        Resolution  & [11, 17] & [6, 10]  & [1, 5] & all\\
       \hline
      {{[}$32^2$, $64^2${]}}&  \textbf{2.97} / 3.01  & \textbf{4.32} / 4.51 & \cellcolor{pastelred!25} 9.23 / \textbf{8.10} & 4.46 / \textbf{4.35}\\
      \hline
      {{[}$64^2$, $96^2${]}}& \textbf{4.41} / 4.57  & \cellcolor{olivine!25} 9.12 / \textbf{9.05} & 16.00 / \textbf{14.80} & 7.26 / \textbf{7.18}\\
      \hline
            {{[}$96^2$, $128^2${]}}& \textbf{6.35} / 6.59  & \cellcolor{olivine!25} \textbf{10.90} / 11.00& 21.10 / \textbf{17.20}  & 7.26 / \textbf{7.18}\\
      \hline
      {{[}$128^2$,{]}}& \cellcolor{darkcyan!25} \textbf{8.89} /  9.16 & 13.96 / \textbf{13.82} & \textbf{33.00} / 33.80 & \textbf{13.40} / 13.60\\

       \hline\hline
       \% Occlusion  & [11, 17] & [6, 10]  & [1, 5]  & all\\
      \hline    
    $>50\%$ & 16.6 / \textbf{15.2} & 16.0 / \textbf{14.8} & \cellcolor{pastelred!25} 32.0 / \textbf{28.1} &19.0 / \textbf{17.4}\\
             \hline
        {{[}$10\%$, $50\%${]}}& \textbf{7.02} / 7.36  &\cellcolor{olivine!25} \textbf{6.88} / 7.00 & 23.7 / \textbf{22.8} & \textbf{8.36} / 8.53\\
       \hline

      $<10\%$ &\cellcolor{darkcyan!25} \textbf{4.91} / 5.22 & \textbf{6.78} / 7.18 & 27.1 / \textbf{24.3} &\textbf{5.59} / 5.80\\          
    
       \hline\hline
       all &  \textbf{6.91} / 6.96 & \textbf{8.04} / 8.12 & 28.00 / \textbf{25.40}   & \textbf{8.21} / 8.28\\

       \hline
    \end{tabular}}
    \caption{ Detection / regression EPE comparison on the COCO validation set with a {common SBL backbone}, with separation according to the number of present joints, input size, and the percentage of occlusions.  Presentation format is detection / regression.  Regression outperforms detection with fewer joints present, smaller input sizes and more occlusion, though this phenomenon is obscured once all the factors are averaged due to the dataset distribution. The shaded blue, green, and red represents easy, medium, and hard samples, respectively.}
    \label{epe_size_kpts}
\end{table}

\begin{figure}[ht]
    \centering
    \includegraphics[width = 0.45\textwidth]{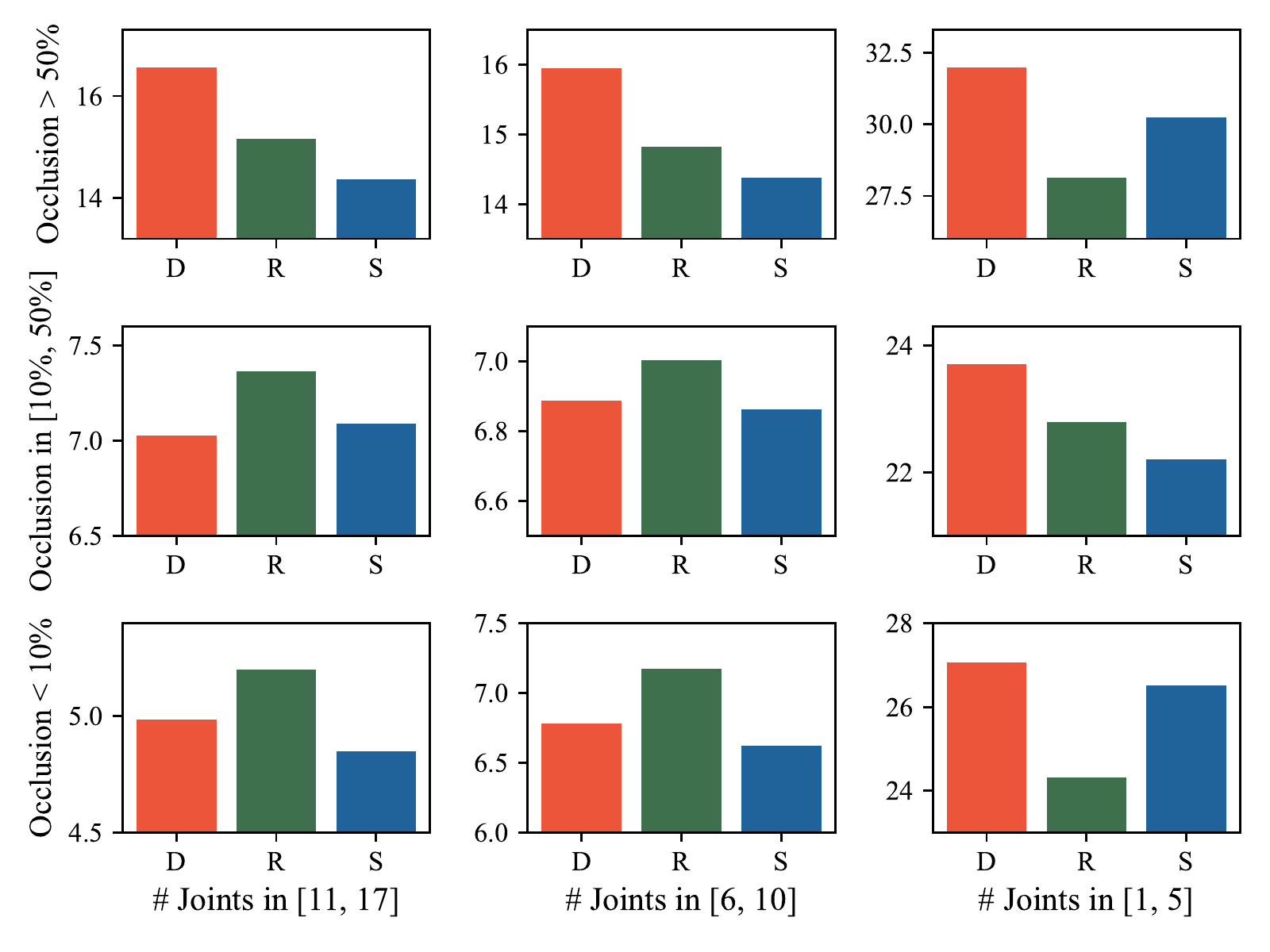}
    \caption{Comparisons of the EPE of our method (S) with detection (D) and regression (R) on the divided subbenchmarks. Our method performs the best, \ie, it has the lowest EPE in 6 of the 9 conditions.
    }
    \label{fig:results_subbenchmark}
\end{figure}

\subsubsection{Comparison Results}
\label{sec:comp_results}

\noindent
\textbf{Training.} We evaluate training by the ascending speed of accuracy with iterations progressing. Fig.~\ref{fig:acc_compare} shows that detection method's initial ascent in average precision (AP) is much faster than integral regression's. Ten epochs of training have already given detection an AP of 80\% of the final value while integral regression takes around sixty epochs. We posit that integral regression's lower training efficiency is a result of the gradient behavior as analyzed in Sec.~\ref{sec:reg_analysis}.

\noindent
\textbf{Inference.} Table~\ref{epe_size_kpts} shows the EPE of integral regression and detection on the divided subbenchmarks. 
Referring to the difficulty label defined in Sec.~\ref{factors}, generally, for easy cases, the detection method performs better; for hard cases, the regression method excels instead; for medium cases, the regression method becomes as competitive as detection. We report more experiments using various backbones on the MSCOCO and MPII dataset in Appendix~E.

Comparing BCIR with detection and integral regression, Fig.~\ref{fig:results_subbenchmark} shows that our method achieves the best result in 6 of the 9 subbenchmarks. Specifically, our method outperforms the regression method in easy cases and the detection method in hard cases, which means it extracts the benefits of both methods.

\begin{figure*}[t]
    \centering
    \includegraphics[width=0.95\linewidth]{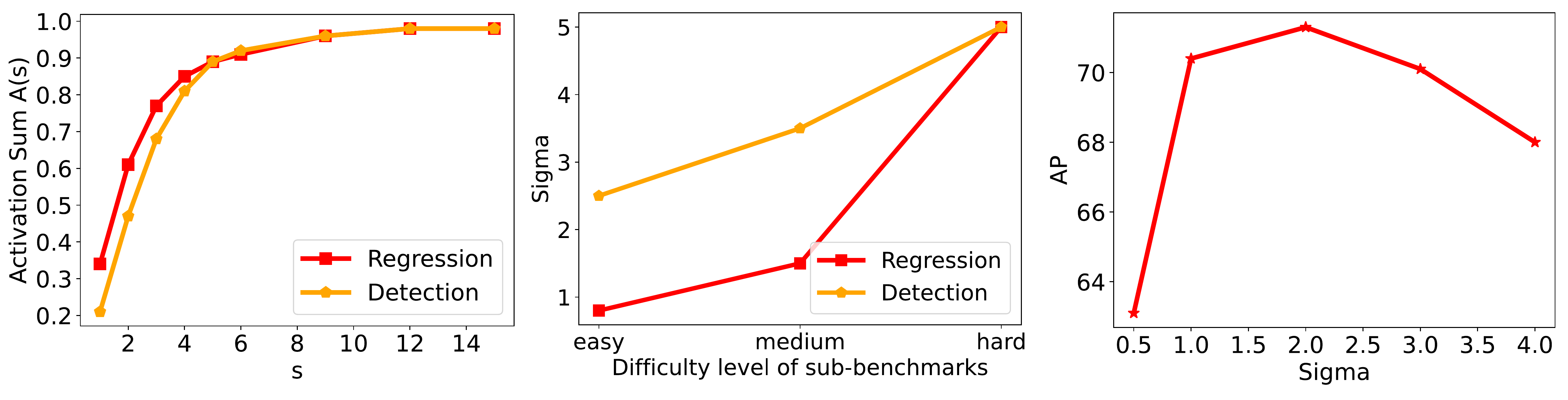}\\[0.5pt] 
    \begin{minipage}[t]{.35\linewidth} 
    \centering {\small (a) Activation sum $A(s)$ of $\tbH$}
    \end{minipage}%
    \begin{minipage}[t]{.3\linewidth} 
    \centering {\small (b) Optimal $\sigma$ of predicted heatmaps}
    \end{minipage}%
    \begin{minipage}[t]{.3\linewidth} 
    \centering {\small (c) AP vs. ground truth $\sigma$}
    \end{minipage}%
    \caption{{\small (a) When $s$ is small, regression is more localized. (b) From hard to easy, the $\sigma$ of detection gradually decreases to 2 but that of regression has no limit. (c) The decrease of $\sigma$ generally improves the performance until it is smaller than 2. 
    }
    }
\label{fig:experimental_support}
\end{figure*}

\subsection{Verification of Localized Heatmaps}
\label{sec:exp_heatmapmodel}
In our localized heatmap model, the assumption of localized activations is reasonable for detection methods as their heatmaps are learned explicitly to match a (localized) ground truth Gaussian. We empirically verify this model by statistically tallying the activations in the normalized heatmap for both detection and integral regression methods based on the easy/medium/hard splits defined in Sec.~\ref{factors}.

Specifically, for each produced heatmap of $64\!\times\!48$, we estimate an activation sum $A$ by summing the normalized heatmap activations within a $2s+1$ square around the ground truth coordinates, $(J_x, J_y)$:
\begin{equation}\label{eq:activation_sum}
    A(s) = \sum_{i=J_x - s}^{J_x+s} \sum_{j=J_y - s}^{J_y+s} \tilde{h}_{ij}.
\end{equation}
The larger $A$ is, the more localized the activations within the region around the ground truth joint, with a maximally achievable $A$ of 1.  Plotting $A$ versus $s$ (see Fig.~\ref{fig:experimental_support}(b)), we observe that a region of $19\times19$, \ie $s=9$, is sufficient to contain 95\% of all non-zero activations for both detection and regression values.  Unsurprisingly, the containing region is smaller for easy samples, \ie, activations on the heatmap are more localized and get progressively larger for the medium and hard samples as the network becomes more uncertain in determining the joint location (see a larger plot in Appendix~F). Specifically, the boost mainly comes from the better performance on easy and medium cases ($\text{AP}_{e/m}$), and the corresponding heatmaps become less localized ($A(s=2)_{e/m}$).

\subsection{Verification that Extremely Localized Heatmaps Degrade Accuracy}
\label{sec:exp_extremelocalize}
Our theory in Sec.~\ref{sec:localization_collapse} suggests that if the standard deviation of the predicted heatmap is extremely small, it will enlarge the distance to ground truth distribution and cause poor performance. 
We verify this for detection methods by applying ground truth Gaussian heatmaps of varying $\sigma$ to train the network. Fig.~\ref{fig:experimental_support}(d) confirms that $\sigma\!=\!2$ appears to be optimal. A smaller $\sigma=1$ degrades the performance slightly, and there is a significant drop in the extreme case when $\sigma\!=\!0.5$. Furthermore, the result shows that the ground truth standard deviation $\sigma$ in Sec.~\ref{sec:localization_collapse} is closest to 2.
To verify the impact of $\sigma$ on regression methods, we propose a simple regularizer, as detailed in Sec.~\ref{sec:regularizer}, to prevent this extremely localized condition. Results in Table~\ref{tab:proposal_performance} show that this regularizer improves the performance.
\subsection{Verification that Integral Regression is More Localized than Detection}
\label{sec:exp_localcomp}
The theory in Sec.~\ref{sec:localization_collapse} is supported by the assumption that integral regression presents a more localized heatmap over detection methods.
To verify this, we compare the activation sum $A(s)$ under the same size $s$ between integral regression and detection.
The plot in Fig.~\ref{fig:experimental_support}(b) indicates that regression method is \emph{more} localized than detection methods. In particular, for a smaller $s$, $A$ is even smaller for detection methods than integral regression methods. We also divide the benchmarks into easy/medium/hard splits and results in~\ref{fig:experimental_support} show that on easy/medium splits, the activation sum difference between the two is larger than that on hard split.

To numerically evaluate the distribution of predicted heatmaps using standard deviation in Sec.~\ref{sec:localization_collapse}, we compare the resulting heatmaps with an idealized Gaussian heatmap of varying standard deviations or $\sigma$, and plot the $\sigma$ which gives the lowest Pearson Chi-square statistic, \ie, the highest similarity (see Fig.~\ref{fig:experimental_support}(c)). The detailed similarities with different standard deviations can be seen in Appendix~G.
As expected, the optimal $\sigma$ decreases as the samples progress from hard to easy; this result is in line with Fig.~\ref{fig:experimental_support}(b) and shows that heatmaps are more localized for easy samples. However, we also observe that the optimal $\sigma$ for regression methods is much smaller for the medium and easy cases. Especially for easy cases, the optimal $\sigma$ is less than 1, which gets far away from the ground truth standard deviation in Sec.~\ref{sec:localization_collapse}.

\begin{table}[ht]
\centering
\scalebox{0.9}{
\begin{tabular}{cllclc}
\toprule
Model & AP & $\text{AP}_{e/m}$ & $\text{AP}_{h}$ & $A(s=2)_{e/m}$ & $A(s=2)_{h}$\\
\midrule
Reg & 67.1  & 70.4 & 44.7  & 0.65 & 0.35\\
\midrule
+$L_{reg}$ & 69.2 (\textbf{+2.1}) & 72.6 (\textbf{+2.2})  & 45.1 & 0.52 (\textbf{-0.13}) & 0.36 \\
\bottomrule
\end{tabular}}

\caption{\small Performance and activation sum of regression baseline and regression baseline with regularizer loss. Compared with regression baseline, the spread of heatmaps is smoother and the results are improved for regression baseline with regularizer loss.}
\label{tab:proposal_performance}
\end{table}

\begin{figure*}[ht]
    \centering
    \includegraphics[width=0.9\textwidth]{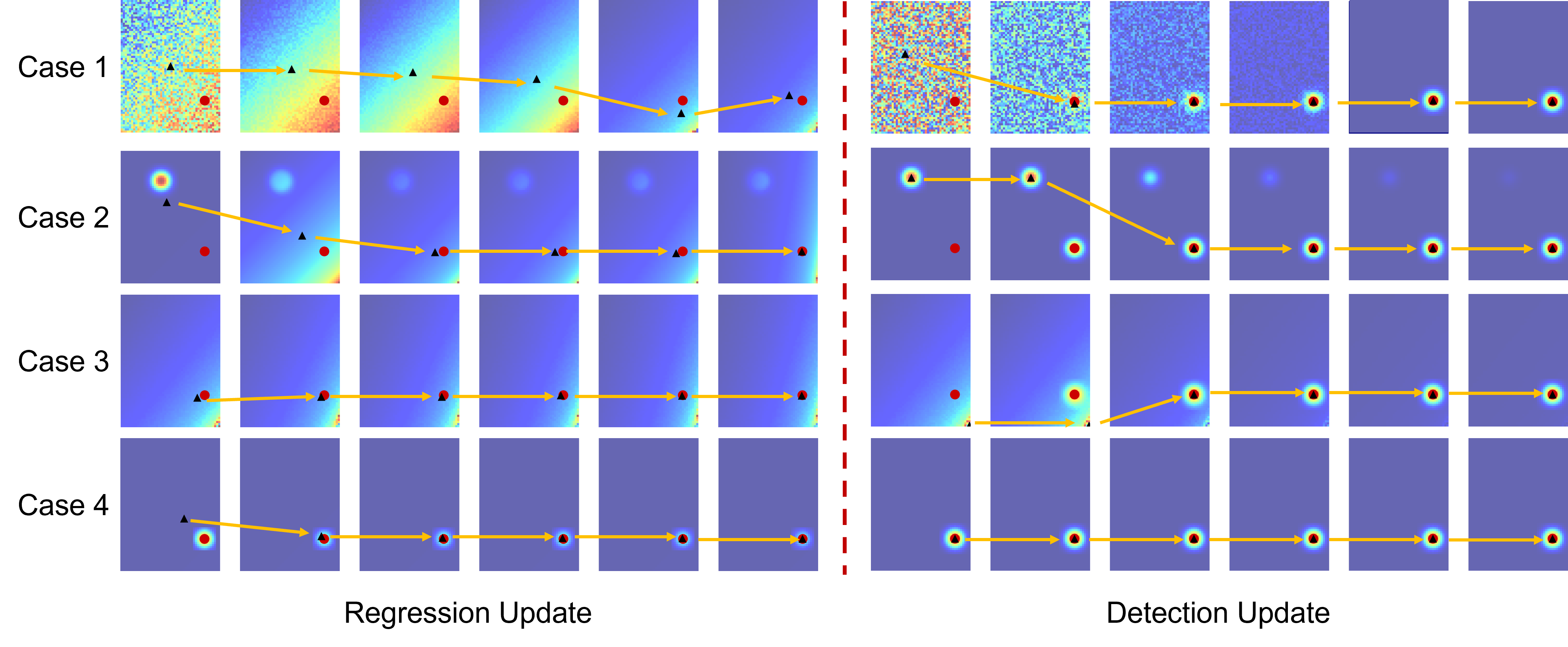}
    \caption{\small Toy example of regression update (left) and detection update (right). 
    The black triangle denotes the prediction (expectation value for regression and max value for detection) while the red dot denotes the ground truth position. The arrow shows the movement of the prediction. 
    }
    \label{fig:rede_toy}
\end{figure*}
\subsection{Verification of Integral Regression Gradients}\label{sec:gradexp}

\noindent
\textbf{Idealized sample.}
We start by considering the case of one sample for a single joint and visualize the progression of a $(64, 48)$ heatmap as it gets updated by Eq.~\ref{eq:h_update}.  Note that while 
$\nabla_{\bp}^n$ can be calculated mathematically by Eq.~\ref{eq:chain_rule}, however we use autograd in Pytorch to obtain the gradients which we verify to be equivalent.  We manually initialize the four cases from Sec.~\ref{sec:reg_analysis} using randomized initialization for case (1), a Gaussian circle centered at some upper left point for case (2), a linear plane at the lower right quadrant for case (3), and a Gaussian circle centered at the ground truth location with $\sigma\!=\!2$ for case (4). The progression of the differently initialized heatmaps can be observed in Fig.~\ref{fig:rede_toy}.

Mathematically, it is clear from Eq.~\ref{eq:taking_expectation} that the resulting joint coordinate location from IPR can align with the ground truth joint even if the underlying heatmap does not follow a localized heatmap model. This is also illustrated in the first three cases of Sec.~\ref{sec:reg_analysis}.  However, we posit that during real-world training, over the thousands of training samples observed over many epochs, the network can only consistently lower the loss if the heatmap $\tilde{\bH}$ is learned to represent $P(\bJ|I)$.  As such, the activation region $\Phi$ will be correctly localized over the corresponding semantic region in the image. Instead, what slows down convergence is the learning of $\Phi$, or rather, the lack of direct guidance to yield a correct $\Phi$ as per detection methods.  For the same four cases, we visualize the updated heatmap of Eq.~\ref{eq:h_update} from the gradients defined by the detection loss in Eq.~\ref{eq:de_backward}. Unlike regression, each case leads consistently to the same localized $\Phi$ centered on the ground truth joint coordinate, regardless of the initialization.

\begin{figure*}[t]
    \begin{minipage}[t]{\linewidth}
    	\centering
    	\includegraphics[width=\linewidth]{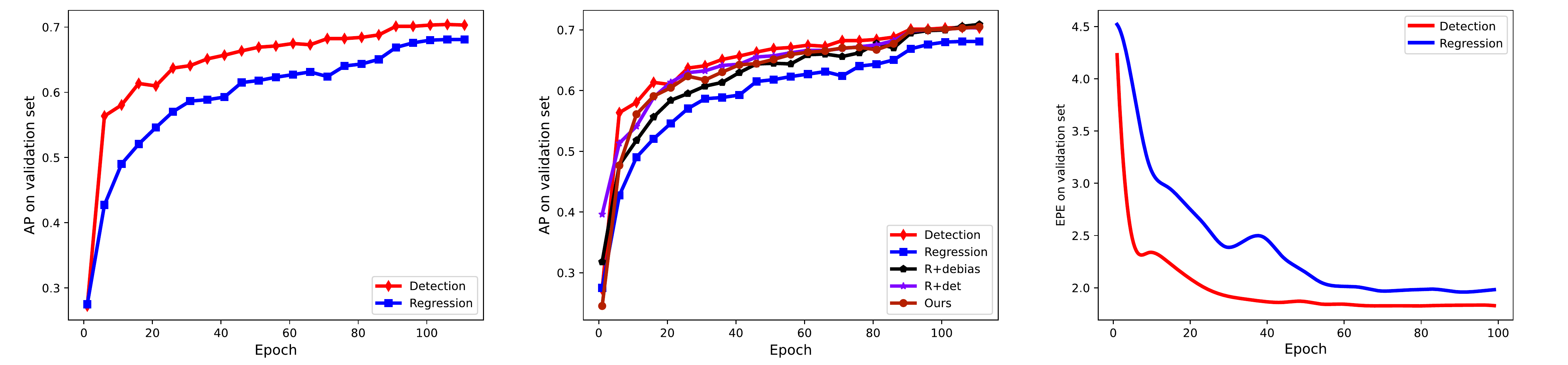}
    \end{minipage}\\
    \begin{minipage}[t]{0.33\linewidth}
    	\centering {\small (a) {Training efficiency of detection vs. regression method} }
    \end{minipage}
    \begin{minipage}[t]{0.33\linewidth}
    	\centering {\small (b) {Comparison of training efficiency among different methods}}
    \end{minipage}
    \begin{minipage}[t]{0.33\linewidth}
    	\centering {\small (c) {Training comparison on RHD}}
    \end{minipage}
    \caption{\small (a) Comparison of the mean Average Precision (AP) on COCO validation set between integral regression and detection. Integral regression takes around six time the number of epochs of detection to reach 80\% of its final value. (b) Influence of each component on the convergence speed on the COCO validation set. Our proposed components accelerates the training speed of regression method so that it approaches the detection method. (c) EPE of detection decreasing faster than regression, especially at the beginning of training.
    }
    \label{fig:acc_compare}\label{fig:training_curve}\label{fig:acc_compare_rhd}
\end{figure*}

\noindent
\textbf{Real world samples.}
The four cases listed in Sec.~\ref{sec:reg_analysis} are not so clearly observable in real-world training as the output is based on the optimization results of batch-wise training on an actual network. Fig.~\ref{fig:acc_compare} helps to verify this, showing the slower convergence of the regression method. 

\begin{figure}[t]
    \centering

    \includegraphics[width=0.45\textwidth]{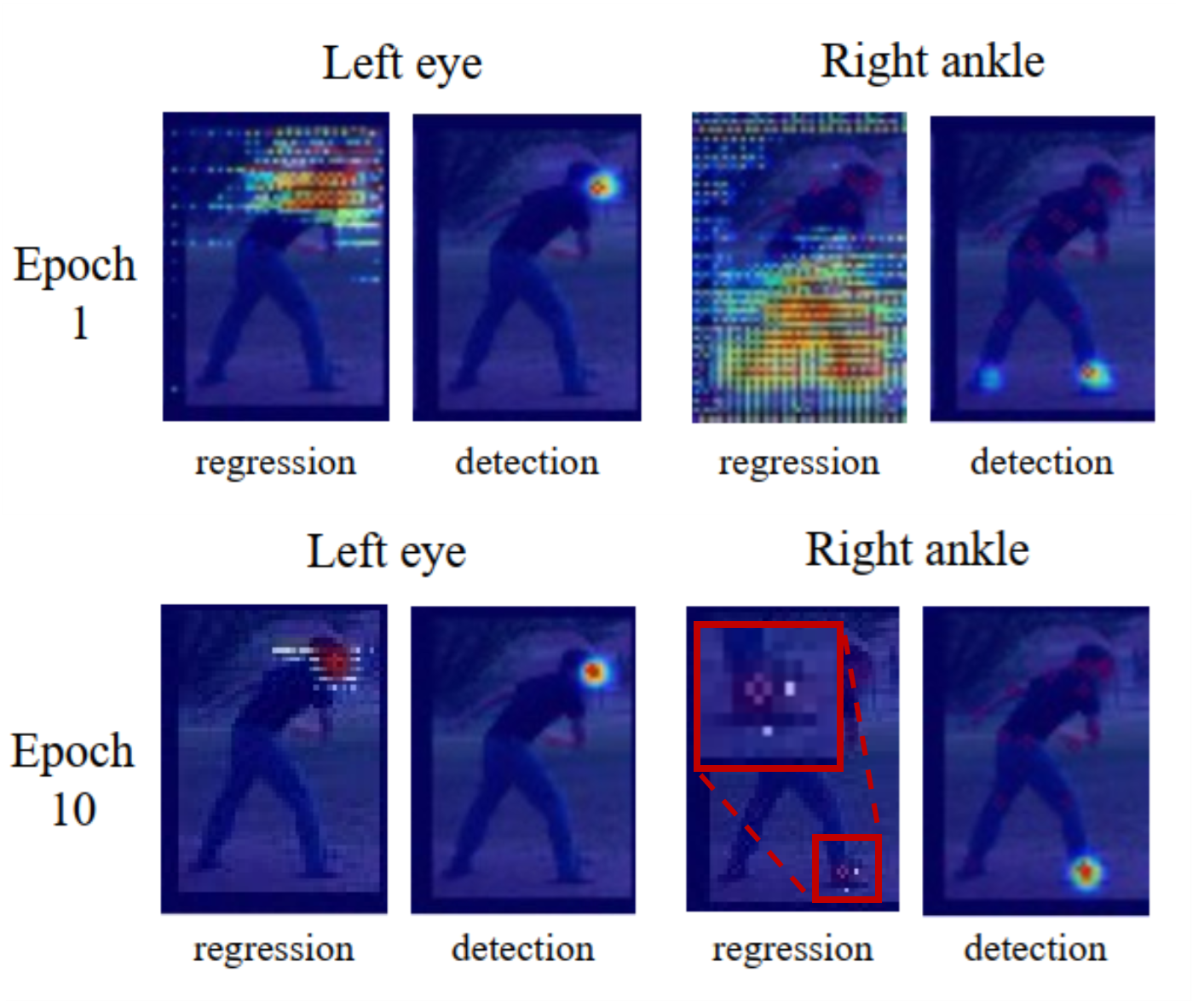}
    \caption{Comparisons of detection and regression in real world training on an image selected from the COCO val set at epoch 1 and 10. Results of detection have already roughly stabilized while regression can only localize to a small extent in epoch 10 with some background pixels still activated. The right ankle is enlarged for regression in the red box for better visualization. More examples are shown in Appendix~H.}
    \label{fig:real_scenario}
\end{figure}

We visualize for one sample from the MSCOCO validation set the heatmap at epochs 1 and 10 to compare the progression of training for detection and integral regression. Fig.~\ref{fig:real_scenario} shows the heatmaps of the ``left eye" and the ``right ankle" joints. For regression, after one epoch, the activations are widespread over a quarter to half of the heatmap.  This aligns roughly to a mix of Case 1 and 3, 
where activations are in the correct quadrant but not yet localized to a small region of support. After 10 epochs of training, predictions for the left eye (which is actually occluded, making it a ``hard'' sample) are localized with several activations of approximately the same value (all shown as red).  For the right ankle (which is clearly present and is an ``easy'' sample), the network has predicted heatmap values that have collapsed spatially and dominate at one or two pixel locations.

In contrast, the detection method presents a well-localized heatmap after only one epoch of training. While there is some activation on the left ankle in epoch 1, this is unlikely to cause errors for the argmax decoding.  After 10 epochs, all activations are centered around the correct right ankle. We provide visualization of full body joints in Appendix~H.

\subsection{Comparison with State-of-the-Art Methods}

\noindent
\textbf{Evaluation on MS COCO}. We compare our method with the top performers of 2D human pose estimation models in Table~\ref{tab:coco_vilid} and Table~\ref{tab:coco_test} on the COCO val and test-dev set. Our method is competitive against state-of-the-art detection-based methods and surpass the performance of regression-based methods by a large margin.

\noindent
\textbf{Evaluation on MPII}. In Table~\ref{tab:mpii}, we also compare our method with state-of-the-art models on the MPII validation set with former regression-based methods, including ~\cite{tompson2015efficient}, DSNT~\cite{nibali2018numerical}, IPR~\cite{sun2018integral}, and detection-based methods, including SBL~\cite{xiao2018simple} and HRNet~\cite{sun2019deep}.

\begin{table*}[t]
\setlength{\tabcolsep}{5mm}
\small
    \centering
    \begin{tabular}{l c l c l l}
    \toprule
        Method & Type & Backbone & Input size & AP(\%)$\uparrow$ & EPE(px)$\downarrow$ \\
    \midrule
        Mask-RCNN~\cite{he2017mask} & D & ResNet-50-FPN & - & 62.9 & -\\
        Hourglass~\cite{newell2016stacked} & D & 8-stage Hourglass & 256 $\times$ 192 & 66.9 & -\\
        CPN~\cite{chen2018cascaded} & D & ResNet-50 & 256 $\times$ 192 & 71.6 & - \\
    \midrule
        IPR~\cite{sun2018integral} & R  & ResNet-101 & 256 $\times$ 256 & 67.2 & 9.98\\
        + BCIR & R & ResNet-101  & 256 $\times$ 256 & \textbf{69.1(+1.9)} & \textbf{9.42(-0.56)}\\
    \midrule
        SBL~\cite{xiao2018simple} & D & ResNet-50 &  256 $\times$ 192 & 70.5 & 9.52\\
        + IPR & R & ResNet-50 & 256 $\times$ 192& 68.2 & 9.63\\
        + BCIR & R & ResNet-50& 256 $\times$ 192& \textbf{71.2(+0.7)} & \textbf{8.93(-0.70)}\\
    \midrule
        SBL~\cite{xiao2018simple} & D & ResNet-152 & 384 $\times$ 288& 73.8 & 8.21 \\
        + IPR & R & ResNet-152 & 384 $\times$ 288& 71.3 & 8.28\\
        + BCIR & R & ResNet-152& 384 $\times$ 288& \textbf{74.4(+0.6)} & \textbf{7.82(-0.39)}\\
    \midrule
        HRNet~\cite{sun2019deep}  & D & HRNet-W32 & 256 $\times$ 192& 75.3 & 7.85\\
        + IPR  & R & HRNet-W32 & 256 $\times$ 192& 72.9 & 8.03\\
        + BCIR & R & HRNet-W32 & 256 $\times$ 192& \textbf{75.8(+0.5)} & \textbf{7.47(-0.38)}\\
    \bottomrule
    \end{tabular}
    \caption{Evaluation of our method competing with state-of-the-art methods on COCO validation set. 'D' and 'R' stand for detection- and regression-based methods, respectively. Our proposed method (+BCIR) outperforms both detection- and regression-based (+IPR) baselines.}
    \label{tab:coco_vilid}
\end{table*}

\begin{table*}[t]
\setlength{\tabcolsep}{3mm}
\small
    \centering
    \begin{tabular}{l c l c c c c c}
    \toprule
        Method & Type & Backbone & Input size & \# Params & GFLOPS & AP (\%)$\uparrow$ & AR (\%)$\uparrow$ \\
    \midrule
        Mask-RCNN~\cite{he2017mask} & D & ResNet-50-FPN & - & - & - & 63.1 & 66.5\\
        CPN~\cite{chen2018cascaded} & D & ResNet-Inception & 384 $\times$ 288 & -& - & 72.1 & 78.5 \\
        RMPE~\cite{fang2017rmpe} & D &  PyraNet & 320 $\times$ 256 & 28.1M & 26.7 & 72.3 & - \\
        SBL~\cite{xiao2018simple} & D & ResNet-152 & 384 $\times$ 288 & 68.6M & 35.6 & 73.7 & 79.0 \\
        HRNet~\cite{sun2019deep} & D & HRNet-W48 & 384 $\times$ 288 & 63.6M & 32.9 & 75.5 & 80.5 \\
        MSPN~\cite{li2019rethinking} & D & 4-stg MSPN & 384 $\times$ 288 & - & - & 76.1 & 81.6 \\
        DARK~\cite{zhang2020distribution} & D & HRNet-W48 &  384 $\times$ 288 & 63.6M & 32.9 & 76.2 & 81.1 \\
        UDP~\cite{huang2020devil} & D & HRNet-W48 &  384 $\times$ 288 & 63.8M & 33.0 & 76.5 & 81.6 \\
    \midrule
        DirectPose~\cite{tian2019directpose} & R & ResNet-101 & - & - & - & 63.3 & - \\
        IPR~\cite{sun2018integral} & R & ResNet-101 & 256 $\times$ 256 & 45.0M & 11.0 & 67.8 & - \\
        RLE~\cite{li2021human} & R & HRNet-W32 & 256$\times$192 & -& 7.1 & 75.7 & 81.3 \\
        PRTR~\cite{li2021pose} & R & HRNet-W32 & 512$\times$384 & 57.2M & 37.8 & 72.1 & 79.4 \\
        Poseur~\cite{mao2022poseur}&R&HRNet-W48&384$\times$288 & - & 33.6&78.8&81.8\\
    \midrule
        BCIR & R & HRNet-W48 & 384 $\times$ 288 & 63.6M & 32.9 & 76.1 & 81.0 \\
    \bottomrule
    \end{tabular}
    \caption{Evaluation of our method competing with state-of-the-art methods on COCO test-dev set. 'D' and 'R' stand for detection- and regression-based methods, respectively. Our proposed method is competitive against state-of-the-art detection-based methods and surpasses the performance of regression-based methods by a large margin.}
    \label{tab:coco_test}
\end{table*}

\begin{table}[t]
\small
\setlength{\tabcolsep}{2mm}
    \centering
    \begin{tabular}{l c l l}
    \toprule
       Method & Type & PCKh@0.5(\%)$\uparrow$ & EPE(px)$\downarrow$ \\
       \midrule
       Tompson et al.~\cite{tompson2015efficient} & R & 80.2 & -\\
        DSNT~\cite{nibali2018numerical} & R & 85.7 & - \\
        \midrule
        IPR~\cite{sun2018integral} & R & 86.5 & - \\
        + BCIR & R & \textbf{87.2(+0.7)} & -\\
        \midrule
        SBL-ResNet50~\cite{xiao2018simple} & D & 87.6 & 20.9 \\
        + IPR & R & 86.2 & 21.5\\
        + BCIR & R &\textbf{87.9(+0.3)} & \textbf{20.3(-0.6)} \\
        \midrule
        SBL-ResNet152~\cite{xiao2018simple} & D & 89.6 & 18.3 \\
        + IPR & R & 87.9 & 19.5\\
        + BCIR & R & \textbf{89.9(+0.3)} & \textbf{17.8(-0.5)}\\
        \midrule
        HRNet-W32~\cite{sun2019deep} & D & 90.4 & 16.6\\
        + IPR & R &88.7 & 18.2\\
        + BCIR & R & \textbf{90.6(+0.2)} & \textbf{16.2(-0.4)} \\
        \bottomrule
    \end{tabular}
    \caption{Comparison on MPII validation set. Our method gains significant improvement on the baselines.}
    \label{tab:mpii}
    \vspace{-4mm}
\end{table}

\begin{table}[t]
\small
\setlength{\tabcolsep}{1.5mm}
    \centering
    \begin{tabular}{l l c c c c}
    \toprule
    Method & $\beta$ & AP & EPE & $\text{EPE}^{\text{H}}$ & $A(s)^{\text{e/m}}$  \\
    \midrule
     SBL~\cite{xiao2018simple} & - & 70.5 & 9.52 & 32.1 & 0.49\\ 
     +IPR & 10 & 68.2 & 9.63 & 28.7 & 0.65\\ 
     +$L_{reg}$ & 10 & 69.5 & 9.43 & 28.4&0.52\\
     +$L_{\text{de}}$, $\lambda\!=\!1$ all epochs & 10 & 70.2 & 9.38 & 28.0 & 0.51\\
     +$L_{\text{de}}$ & 10 & 70.4 & 9.31 & 27.8 &0.52\\
     +de-bias& 10 & 71.0 & 9.15 & 27.1 & 0.55\\
     +BCIR ($L_{\text{de}}$+de-bias)& 1 & 70.7 & 9.02 & 26.1 & 0.53\\
     +BCIR ($L_{\text{de}}$+de-bias)& 20 & 70.3 & 9.32 & 30.7 & 0.62\\
     \midrule
     +BCIR ($L_{\text{de}}$+de-bias) & 10 & \textbf{71.2} & \textbf{8.93} & \textbf{25.6} & 0.53\\
     \bottomrule
    \end{tabular} 
    \caption{Evaluation of each component of our method on COCO validation set. $\text{EPE}^{\text{H}}$ denotes EPE on `hard' samples. For $L_{\text{de}}$, $\lambda(t)\!=\!1$ for $T_o\!=\!120$ unless otherwise indicated. Our proposed components improve the performance with respect to the baseline, especially on these hard samples.  A combination with $\beta\!=\!10$ is optimal.}
    \label{tab:ablation}
\end{table}

\subsection{Ablation Study}
\label{sec:ablation_study}

Our method consists of two main components: a bias compensation and a regularization term. The regularization term is initially a simple regularizer (See Sec.~\ref{sec:regularizer}) and further developed to a concrete Gaussian prior with a fixed standard deviation. This subsection features ablation studies to demonstrate the effects of each component.

\noindent
\textbf{Effects of components}.  The baselines are the detection-based and regression-based (+IPR) methods; we compare them with the models combined with detection loss or bias compensation.  Addition, we include variants with stronger regularization terms and when the detection loss is applied for all epochs. Fig.~\ref{fig:training_curve} shows the average precision on the validation set after training specific epochs. Each component speeds up the training. Meanwhile, Table~\ref{tab:ablation} shows that each component contributes to the higher performance, and $L_{\text{de}}$ outperforms $L_{reg}$.

\noindent
\textbf{Selection of hyperparamter $\beta$}. As illustrated above, a small $\beta$ leads to a large bias, and a large $\beta$ makes back propagation difficult. We evaluate the different selection of the $\beta$ value in Table~\ref{tab:ablation} and choose $\beta\!=\!10$ as our optimal value.

\subsection{Verification on Hand Pose Estimation}

To verify whether our findings on hand pose estimation, we extend our analysis and experiments to RHD, a commonly used hand pose dataset. Our experiments showed that in the hand pose estimation task, integral regression (or 2.5D regression in hand pose estimation) also trains slower than detection, and integral regression performs better in occluded conditions.

\noindent
\textbf{Training comparison.}
We compare the training speed of detection and integral regression using SBL on the RHD dataset in Fig.~\ref{fig:acc_compare_rhd}. Similar to the human pose estimation task, detection learns faster than integral regression on the hand pose estimation task.

\noindent
\textbf{Inference comparison} Similar to human pose estimation, we need to divide the evaluated benchmark RHD according to their levels of difficulty. Since the input size has the same scale and all the hand keypoints are present, we can only use occlusion as the criterion. Specifically, we consider the visible keypoints as ''easy'' cases and occluded keypoints as ''hard'' cases. Since the annotators did not provide whether a keypoint is occluded or not, we roughly determine the visibility using the depthmap and the depth information of a keypoint as follows:
\begin{equation}\label{eq:visibility}
    v = 
    \begin{cases}
    1, \,\,\,\,\,\,|D(u, v) - z| < \delta \\
    0, \,\,\,\,\,\,|D(u, v) - z| \geq \delta
    \end{cases},
\end{equation}

\noindent
where $(u,v)$ is the projected $(x,y)$ location of this keypoint, $D(u,v)$ is the depth value of location $(u,v)$ on the depthmap, and $z$ is the depth of this keypoint. The rationale is that if the depth of some joint is much deeper than its surface depth (from depth), we consider it as occluded. The hyperparameter $\delta$ means the thickness of the keypoint combined with errors caused by machine; from observation, we set it to 3mm. We report the EPE to evaluate the performance on the divided samples in Table~\ref{tab:divided_RHD}. When both models are well trained, the overall performance mainly comes from the easy cases ($v\!=\!1$). We also present a checkpoint for detection models at the 30\emph{th} epoch when the overall performance is roughly the same. Results showed that IPR performs better in the hard cases ($v\!=\!0$) by 0.1 even though the general performance is slightly worse.

\begin{table}[t]
    \centering
    \begin{tabular}{lccc}
    \toprule
    Method & avg & $v\!=\!1$ & $v\!=\!0$  \\
    \midrule
    IPR  & 2.02 &  1.84 & 2.45 \\
    \midrule
    Detection & 1.87 & 1.58 & 2.44 \\
    \midrule
    Detection (30 epochs) & 1.99 & 1.78 & 2.55\\
    \bottomrule
    \end{tabular}
    \caption{Comparisons of EPE on the RHD test set. avg is the overall performance and $v$ is the visibility of the samples. The well-trained models mainly have different performance in visible cases and under the similar overall performance, IPR performs much better in occluded cases.}
    \label{tab:divided_RHD}
\end{table}

\begin{table}[t]
\small
\setlength{\tabcolsep}{2mm}
    \centering
    \begin{tabular}{l c l l}
    \toprule
       Method & Type & AUC(\%)$\uparrow$ & EPE(px/mm)$\downarrow$ \\
       \midrule
      Z\&B~\cite{zimmermann2017learning} & D & 72.0/67.5 & 9.14/30.4\\
        Cai~\cite{cai2018weakly} & D & -/88.7 & -/-\\
        \midrule
        2.5D regression~\cite{iqbal2018hand} & R & 84.4 / 93.0 & 4.76 / 14.3\\
        + BCIR & R & \textbf{{85.8 / 93.6}} & \textbf{4.34 / 13.5}\\
        \bottomrule
    \end{tabular}
    \caption{Comparison on the RHD test set of 2D/3D AUC and EPE for hand pose estimation.  Our proposed method outperforms the baseline and two detection-based methods.}
    \label{tab:rhd}
\end{table}

\noindent \textbf{Comparison with Hand Pose Methods} We compare our method with a regression-based method, 2.5D regression~\cite{iqbal2018hand} and two detection-based methods~\cite{zimmermann2017learning,cai2018weakly} in Table~\ref{tab:rhd}. BCIR is built on top of 2.5D regression method and achieves better AUC and EPE on both 2D and 3D space.

\section{Conclusion}

This paper presented a comprehensive analysis of the commonly used heatmap representation in the pose estimation task.  It also introduces the first systematic comparison between the two different methods of heatmap decoding and supervision: detection versus integral regression.
For integral regression, the combination of softmax and expectation operations induces a bias and results in extremely localized heatmaps during training. 
These localized heatmaps deviate from true distributions of keypoints and deteriorate overall accuracy. In addition, the derivatives of heatmaps with respect to integral regression loss are less informative than detection loss and make the training progress less efficient.

Based on these observations, we propose a bias-compensated regression method along with a Gaussian prior loss to speed up the training and improve the performance on human and hand pose estimation tasks. Our method functions as an alternative to detection-based methods to achieve high accuracy of coordinates via the heatmap in an end-to-end way. Especially for tasks that use coordinates as input, decoding directly from the heatmap with a non-differentiable argmax may not be possible. By adding a spatial prior and compensating for the bias, the predicted distribution gets closer to the true one compared with the original soft-argmax, giving better input to the subsequent tasks. We hope our method can work as a better solution in decoding the heatmaps to coordinates when the heatmaps are not supposed to be strictly one-hot.

\ifCLASSOPTIONcompsoc
  \section*{Acknowledgments}
\else
  \section*{Acknowledgment}
\fi

This research / project is supported by the Ministry of Education, Singapore, under its MOE Academic Research Fund Tier 2 (STEM RIE2025 MOE-T2EP20220-0015).

\ifCLASSOPTIONcaptionsoff
  \newpage
\fi



%

\bibliographystyle{IEEEtran}
\bibliography{IEEEbib}




%
\begin{IEEEbiography}[{\includegraphics[width=1in,height=1.2in,clip,keepaspectratio]{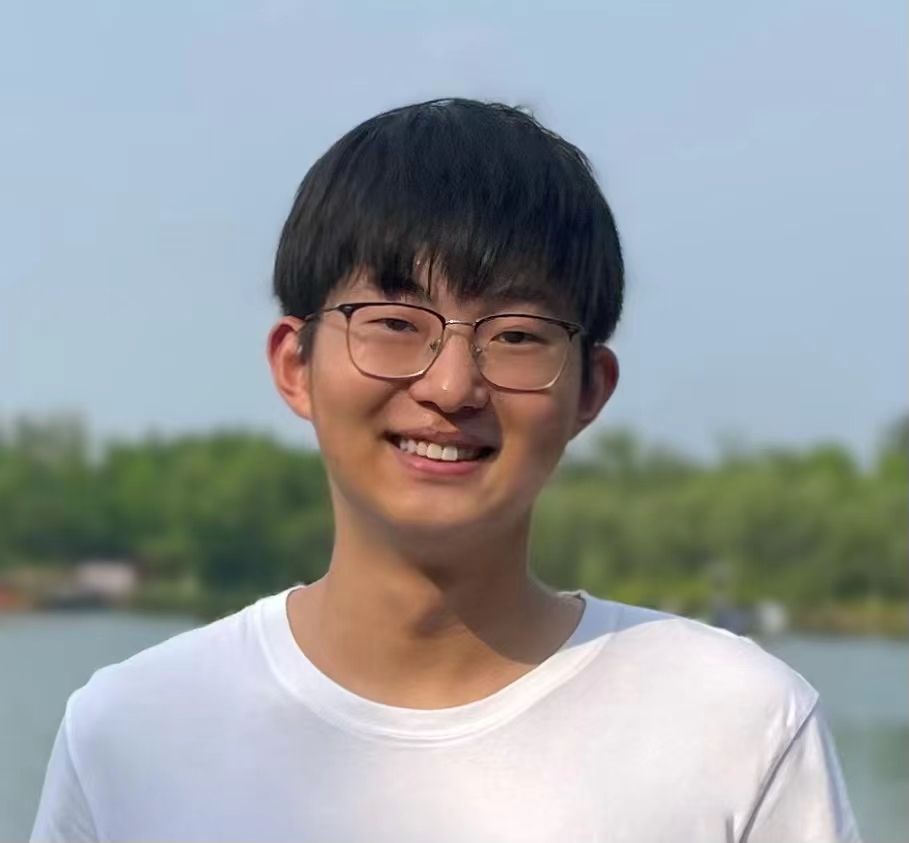}}]{Kerui Gu} is a Ph.D. student in the School of Computing at the National University of Singapore. He received Bachelor's degree from Shanghai Jiao Tong University in 2016. His research interest includes 2D/3D human pose estimation.
\vspace{-2em}
\end{IEEEbiography}

\begin{IEEEbiography}[{\includegraphics[width=1in,height=1in,clip,keepaspectratio]{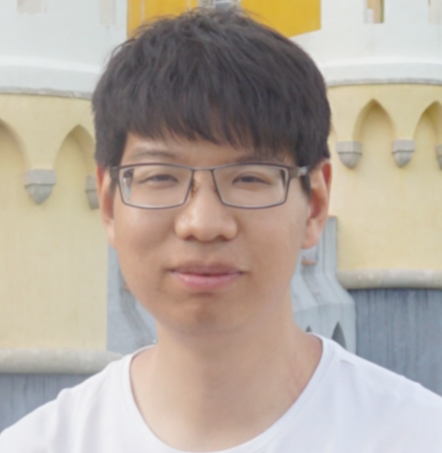}}]{Linlin Yang} is a Ph.D. candidate at University of Bonn and a research associate at National University of Singapore. His research focuses on pose estimation and self-/semi- supervised learning.
\vspace{-2em}
\end{IEEEbiography}

\begin{IEEEbiography}[{\includegraphics[width=1in,height=1.25in,clip,keepaspectratio]{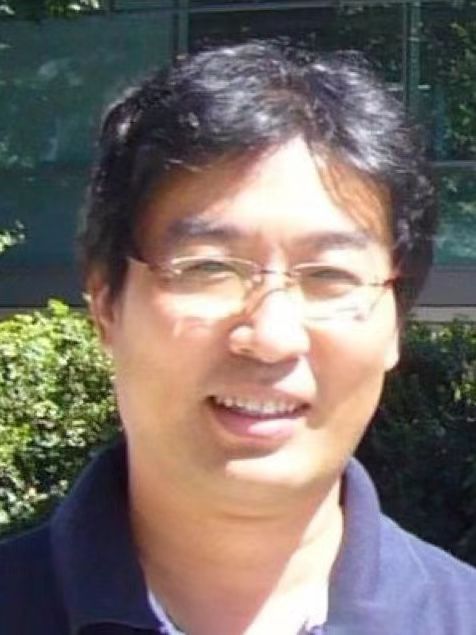}}]{Michael Bi Mi} received the Ph.D. degree from National University of Singapore in 1996. He is currently a research scientist in Huawei International, Singapore. Before that, he had been working for 19 years with Panasonic Singapore Laboratories, where he served as a General Manager of System Architecture Group. His research interests include deep learning, model compression,  computer vision, image and video processing, etc. 
\end{IEEEbiography}

\begin{IEEEbiography}[{\includegraphics[width=1in,height=1.25in,clip,keepaspectratio]{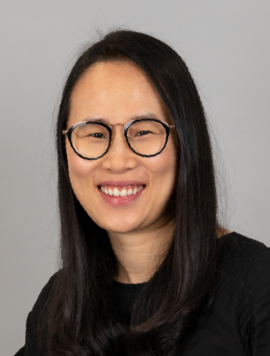}}]{Angela Yao} is a Dean's Chair Assistant Professor in the School of Computing at the National University of Singapore. She received a PhD from ETH Zurich and a BASc from the University of Toronto. Angela leads the Computer Vision and Machine Learning group, with a special focus on vision-based human motion analysis. She is the recipient of the German Pattern Recognition (DAGM) award (2018) and Singapore’s National Research Foundation's Fellowship in Artificial Intelligence (2019).
\end{IEEEbiography}








\clearpage
\input{appendix.tex}
\end{document}

%% file: appendix.tex
%


%

\makeatletter
\setlength{\abovecaptionskip}{6pt}   
\setlength{\belowcaptionskip}{6pt}   
\long\def\@makecaption#1#2{%
  \vskip\abovecaptionskip
  \sbox\@tempboxa{#1: #2}%
  \ifdim \wd\@tempboxa >\hsize
    #1: #2\par
  \else
    \global \@minipagefalse
    \hb@xt@\hsize{\box\@tempboxa\hfil}%
  \fi
  \vskip\belowcaptionskip}
\makeatother

%
\ifCLASSOPTIONcompsoc
\else
  \usepackage{cite}
\fi
%

%
\ifCLASSINFOpdf
\else
\fi
\title{Appendix of Bias-Compensated Integral Regression for Human Pose Estimation}

\maketitle

\appendices
\section{First Derivative of $D_B(\bJ_{\text{gt}}, \hbJ, \hat{\sigma}^2)$}
\label{app:first_derivative_of_db}

The first derivative of $D_B(\bJ_{\text{gt}}, \hbJ, \hat{\sigma}^2)$ can be derived as follows:
\begin{equation}
    D_B^{'}(\hat{\sigma}^2) = (\frac{-\frac{\sigma_{\text{true}}^2}{\hat{\sigma}^4}+\frac{1}{\sigma_{\text{true}}^2}}{(\frac{\sigma_{\text{true}}^2}{\hat{\sigma}^2}+\frac{\hat{\sigma}^2}{\sigma_{\text{true}}^2}+2)})-\frac{1}{4}(\frac{\bmu - \bmu_{\text{true}}}{\sigma_{\text{true}}^2+\hat{\sigma}^2})^2
\end{equation}
Setting $D_B^{'}$ to 0 and we can obtain Eq.~14 in the main paper.

\section{Derivation of Bias}
\label{app:derivation_of_bias}

As defined in the main paper, we obtain the normalized heatmap using the soft-argmax function as follows:

\begin{equation}\label{eq:softmax}
    \Tilde{\bH}(\bp) = \frac{{\rm exp}(\beta\cdot {\bH}(\bp))}{\sum_{\bp^{'}\in \Omega}{\rm exp}(\beta \cdot {\bH}(\mathbf{p}^{'}))}, \quad \beta > 0,
\end{equation}

\noindent where $\bH(\bp)$ is the heatmap output of the network and indexed by pixel $\bp$ over the range of pixels $\Omega$.  For convenience, we further define a variable $C$ as the denominator of Eq.~\eqref{eq:softmax}:

\begin{equation}
    C = {\sum_{\bp{'}\in \Omega}{\rm exp}(\beta \cdot {\bH}(\mathbf{p}{'}))}.
\end{equation}

We can further partition the heatmap's pixels $\Omega$ into four sections $\{\Omega_1, \Omega_2, \Omega_3, \Omega_4\}$, as visualized in Fig.~2 in the main paper.   $\Omega_1$ is defined such that the true joint location $(x_o, y_o)$ is the expected value and the center of the section. The key assumption that we make in our work is that the heatmap support for the true joint location $(x_o, y_o)$ is well localized and fully contained within $\Omega_1$ in $\bH$.  As such, the sections $\Omega_2$ to $\Omega_4$ contain only zero or near-zero elements so we can approximate Eq.~\eqref{eq:softmax} for the four sections as follows:

\begin{align}
 \tilde{\mathbf{H}}(\mathbf{p}) \approx \begin{cases}
 \frac{1}{C} \cdot \exp(\beta_k \mathbf{H}(\mathbf{p})) \quad & \text{for } \mathbf{p} \in \Omega_1\\
 \frac{1}{C} & \text{for } \mathbf{p} \in \{\Omega_2,\Omega_3,\Omega_4\},\\
 \end{cases}
\end{align}
where the normalized heatmap value approximates to $1/C$ for $\bp \in \{\Omega_2,\Omega_3,\Omega_4\}$ since the exponential of a zero in the numerator is simply 1.

The (biased) joint location $\bJ^r(x_r, y_r)$ is defined as the expected value of the entire heatmap, which can be further decomposed into the four sections: 

\begin{align}
    \mathbf{J}^r & = \sum\limits_{\mathbf{p} \in \Omega}\Tilde{\mathbf{H}}(\mathbf{p})\cdot\mathbf{p} \\
    & = \sum\limits_{\mathbf{p} \in \Omega_1}\Tilde{\mathbf{H}}(\mathbf{p})\cdot\mathbf{p} \,\,\,\, +  \!\!
    \sum\limits_{\mathbf{p} \in \Omega_2,\Omega_3,\Omega_4} \!\! \Tilde{\mathbf{H}}(\mathbf{p})\cdot\mathbf{p}.
\end{align}

We can also view $\bJ^r = (x_r, y_r)$ as a weighted sum of the expected location of each section:

\begin{equation}
    \begin{aligned}
        \mathbf{J}^r = w_1\mathbf{J}_1 + w_2\mathbf{J}_2 + w_3\mathbf{J}_3+w_4\mathbf{J}_4, \\
        \text{where } 
        w_k = \sum_{\bp\in \Omega_k} \Tilde{\bH}(\bp), \quad \text{for } k=1,2,3,4
    \end{aligned}
    \label{eq: gaussianmap}
\end{equation}

\noindent
where $\bJ_1 = (x_o, y_o)$, $\mathbf{J}_2 = (x_o, y_o+\frac{w}{2})$, $\mathbf{J}_3 = (x_o+\frac{h}{2}, w/2)$, and $\mathbf{J}_4 = (x_o+\frac{h}{2}, y_o+\frac{w}{2})$.  Due to the symmetry of each region, we can also represent the weights $w_2$ to $w_4$ as an expression of $w, h$ and $C$.

\begin{equation}
\begin{aligned}
    w_2 & = \frac{1}{C} \cdot {2x_o(w-2y_o)}, \\
    w_3 & = \frac{1}{C} \cdot {2(h-2x_o)y_o}, \\
    w_4 & = \frac{1}{C} \cdot {(h-2x_o)(w-2y_o)}. \\
\end{aligned}
\label{eq:weights}
\end{equation}

\noindent We can reformulate Eq.~\eqref{eq: gaussianmap} in matrix format:

\begin{equation}
    \begin{bmatrix}
    x_r \\
    y_r
    \end{bmatrix} = 
    \begin{bmatrix}
    w_1x_o + w_2x_o+w_3(x_o+\frac{h}{2})+w_4(x_o+\frac{h}{2})\\ 
    w_1y_o + w_2(y_o+\frac{w}{2})+w_3y_o+w_4(y_o+\frac{w}{2})
    \end{bmatrix}.
    \label{eq:x_rwithweights}
\end{equation}

\noindent Substituting the weights from Eq.~\eqref{eq:weights} into Eq.~\eqref{eq:x_rwithweights} and with the knowledge that $w_1 = 1\!-\!w_2\!-\!w_3\!-\!w_4$, we arrive at the following linear equation:

\begin{equation}
\bJ^r = 
    \begin{bmatrix}
    x_r \\
    y_r
    \end{bmatrix} = 
    \begin{bmatrix}
    (1-\frac{hw}{C})x_o + \frac{hw}{C}\frac{h}{2} \\ 
    (1-\frac{hw}{C})y_o + \frac{hw}{C}\frac{w}{2}
    \end{bmatrix}.
    \label{eq:linear-bias-supp}
\end{equation}

\begin{table}[t]
\small
    \centering
    \begin{tabular}{l c c c c}
    \toprule
      De/Re  & few kpts & some kpts  & many kpts &all\\
      \midrule
      many occ & 28.5/\textbf{26.6} & 14.1/\textbf{13.3} & 14.2/\textbf{14.7} &16.8/\textbf{16.6}\\
      \midrule
      some occ & 22.2/\textbf{20.5} & \textbf{6.72}/6.93 & \textbf{7.18}/7.20 & 8.27/\textbf{8.20}\\
      \midrule
      few occ & 26.9/\textbf{26.4} & \textbf{7.19}/7.38 & \textbf{5.12}/5.29 &\textbf{5.78}/5.95\\
      \midrule
      all & 25.8/\textbf{24.1} & \textbf{7.98}/8.10 & \textbf{6.74}/6.93 & \textbf{8.09}/8.14\\
      \bottomrule
    \end{tabular}
    \caption{HRNet - Comparisons of EPE on the COCO validation set. De and Re refers to detection and regression method, respectively.}
    \label{epe_occ_kpts}
\end{table}

\begin{table}[t]
\footnotesize
    \centering
    \begin{tabular}{l c c c c}
    \toprule
         & [0,5] & [6, 10]  & [11, 17] &all\\
       \midrule
       {{[}$32^2$, $64^2${]}} & 401 & 794 & 1218 & 2413\\
       \midrule
       {{[}$64^2$, $96^2${]}} & 174 & 327 & 747 & 1248\\
       \midrule
       {{[}$96^2$, $128^2${]}} & 103 & 187 & 492 & 782\\
       \midrule
       {{[}$128^2$,{]}} & 256 & 484 & 1169 & 1909\\
       \midrule
       all & 934 & 1792 & 3626 & 6352\\
       \bottomrule
    \end{tabular}
    \caption{Number of person instances when separating the benchmarks according to the number of present joints (column) and the input size (row).} 
    \label{num_size_kpts}
\end{table}

\begin{table}[t]
\footnotesize
    \centering
    \begin{tabular}{l c c c c}
    \toprule
          & [0,5] & [6, 10]  & [11, 17] &all\\
       \midrule
        $>50\%$    & 167 & 149 & 52 & 368\\
       \midrule
        {{[}10\%, 50\%{]}} & 182 & 584 & 602 & 1368\\
       \midrule
       $<10\%$ & 585 & 1059 & 2972 & 4616\\
       \midrule
       all & 934 & 1792 & 3626 & 6352\\
       \bottomrule
    \end{tabular}
    \caption{Number of person instances when separating the benchmarks according to number of present joints (column) and percentage of occlusions (row).}
    \label{num_occ_kpts}
\end{table}

\begin{table}[t]
\footnotesize
    \centering
    \begin{tabular}{l c c c}
    \toprule
        D/R/S &  [0,5] & [6, 10]  & [11, 17] \\
       \midrule
        $>50\%$    &  32.0 / \textbf{28.1}/28.2 & 23.7 / 22.8/\textbf{22.2} &  27.1 / 24.3/\textbf{23.8}\\
       \midrule
       {{[}10\%, 50\%{]}} & 16.0 / 14.8/ \textbf{14.4} & 6.88/ 7.00/\textbf{6.86} & 6.78 / 7.18/\textbf{6.62}\\
       \midrule
       $<10\%$ & 16.6 / 15.2/\textbf{14.3} & 8.36 / 8.53/\textbf{7.08} & 4.91 / 5.22/\textbf{4.84} \\
       \bottomrule
    \end{tabular}
    \caption{Comparison of EPE of our method with detection and regression-based method on subbenchmarks divided according to our proposed method on the COCO validation set.} 
    \label{number_figure_5}
\end{table}

Even though we began our derivation with $(x_o,y_o)$ located in $\Omega_1$ which is in the upper left quadrant, Eq.~\eqref{eq:linear-bias-supp} is equally applicable when $(x_o,y_o)$ is located in the other three quadrants. When we look at Eq.~\eqref{eq:linear-bias-supp}, if $x_o<\frac{h}{2}$, then  $x_r>x_o$, which pushes the coordinate toward the center. If $x_o>\frac{h}{2}$, then $x_r<x_o$, which also makes the prediction closer to the center. $y_o$ is the same as $x_o$. Therefore, this equation is applicable to all quadrants.

We can predict $\bJ^o$ from $\bJ^r$ in closed form as follows:

\begin{equation}
    \bJ^{ro} =
    \begin{bmatrix}
    x_0 \\
    y_0
    \end{bmatrix} = 
    \begin{bmatrix}
    \frac{C}{C-hw}x_r - \frac{hw^2}{2(C-hw)} \\ 
    \frac{C}{C-hw}y_r - \frac{h^2w}{2(C-hw)}
    \end{bmatrix},
    \label{eq:linear-bias-reverse}
\end{equation}

\noindent which is the result in the main paper.

\section{Derivation of Eq.~23 in the main paper}
\label{sec:chain_rule_appendix}

Eq.~23 in the main paper is obtained by the chain rule. Therefore, we list the value of each term as follows

\begin{multline}
    \frac{\partial L_\text{re}}{\partial \hat{\bJ}} =     \begin{bmatrix}
        s(\hat{J}_x - J_x)\\
        s(\hat{J}_y - J_y)
    \end{bmatrix},\,\,\,
\frac{\partial \hat{\bJ}}{\partial \tilde{h}_{\bp}} = 
(i+j),\,\,\, \\
\frac{\partial \tilde{h}_{\bp}}{\partial \hat{{h_{\bq}}}}=\begin{cases}
    \beta \tilde{h}_{\bq} (1 - \tilde{h}_{\bq}), \,\,\bq = \bp \\
    -\tilde{h}_{\bp}\tilde{h}_{\bq}, \,\,\,\quad \bq\neq \bp
    \end{cases},
\end{multline}

\noindent
where $\bq = (u ,v)$ can be any pixel in $\Omega$.
Combining the terms, gradients at $\bp$ can be calculated by

\begin{multline}
\label{eq:chain_rule_long}
    \frac{\partial L_\text{re}}{\partial \hat{h}_{\bp}} = 
    \beta ( (\hat{J}_x - J_x)i+ (\hat{J}_y - J_y)j) \tilde{h}_{\bp} (1 - \tilde{h}_{\bp}) \\ 
    - \sum_{\substack{\bq \in \Omega, \\ \bq \neq \bp}}\beta ( (\hat{J}_x - J_x)u+ (\hat{J}_y - J_y)v)\tilde{h}_{\bq}\tilde{h}_{\bp}.
\end{multline}

\noindent

Eq.~23 of the main paper can be obtained by rearranging the terms of Eq.~\ref{eq:chain_rule_long}.

\section{Derivation of Eq.~24 in the main paper}
\label{sec:derivation_taylor}

Let us suppose $\bH_n=g(\bw_n|I)$ where $\bH_n$ is the heatmap of a given image $I$ at the $n$-th iteration, $g(\bw_n)$ is the function of generating the heatmap, \ie the backbone network, with weights at the $n$-th iteration. According to a Taylor expansion, we arrive at
\begin{multline}
\label{eq:taylor_ori}
    g(\bw_{n+1}) = g(\bw_n) + \frac{g^{'}(\bw_n)}{1!}(\bw_{n+1}-\bw_n)
    \\+\frac{g^{''}(\bw_n)}{2!}(\bw_{n+1}-\bw_n)^2+\ldots,
\end{multline}
where $\bw_{n+1}\!-\!\bw_n\!\!=\!-\alpha\nabla g(\bw)\!=\!-\alpha\frac{\partial L}{\partial \bH}g^{'}(\bw_n)$ according to gradient descent, and $(\bw_{n+1}\!\!-\!\!\bw_n)^i$ is negligible compared with $(\bw_{n+1}\!-\!\bw_n)$ when $i\!>\!1$. Therefore, Eq.~\ref{eq:taylor_ori} can be rewritten as
\begin{multline}\label{taylor_long}
    \bH_{n+1}=g(\bw_{n+1}) \approx g(\bw_n) + g^{'}(\bw_n)(\bw_{n+1}-\bw_n)\\=\bH_n-\alpha\frac{\partial L}{\partial \bH}|| g^{'}(\bw_n)||^2.
\end{multline}
Therefore, for each pixel $h_{\bp}$ in $\bH$, the update rule is
\begin{equation}
h_{\bp}^{n+1} \approx h_{\bp}^{n} - \gamma\nabla_{\bp}^n,
\end{equation}
where $\gamma$ represents $\alpha \|g^{'}(\bw_n)\|$, $h_{\bp}^n$ and $\nabla_{\bp}^n$ denote heatmap and gradient value located at $\bp$ in the heatmap at the $n$th iteration. $h_{\bp}^0$ denotes the original heatmap.

\section{Experiment Details on Subbenchmarks}
\label{app:comp_subbenchmark}

We report the experiment results of the performance of HRNet~\cite{sun2019deep} on different subbenchmarks in Table~\ref{epe_occ_kpts}.

We report the number of person instances in each subbenchmarks divided by the proposed factors on the COCO~\cite{lin2014microsoft} validation set in Table~\ref{num_size_kpts} and Table~\ref{num_occ_kpts}.

We also report the detailed EPE of our method on the divided sub benchmarks in Table~\ref{number_figure_5} to support the results in Fig.~5 of the main paper.


\section{Extension of Fig.~5(a) in the main paper}
\label{app:as_plot_cases}

\begin{figure}[ht]
    \centering
    \includegraphics[width=0.45\textwidth]{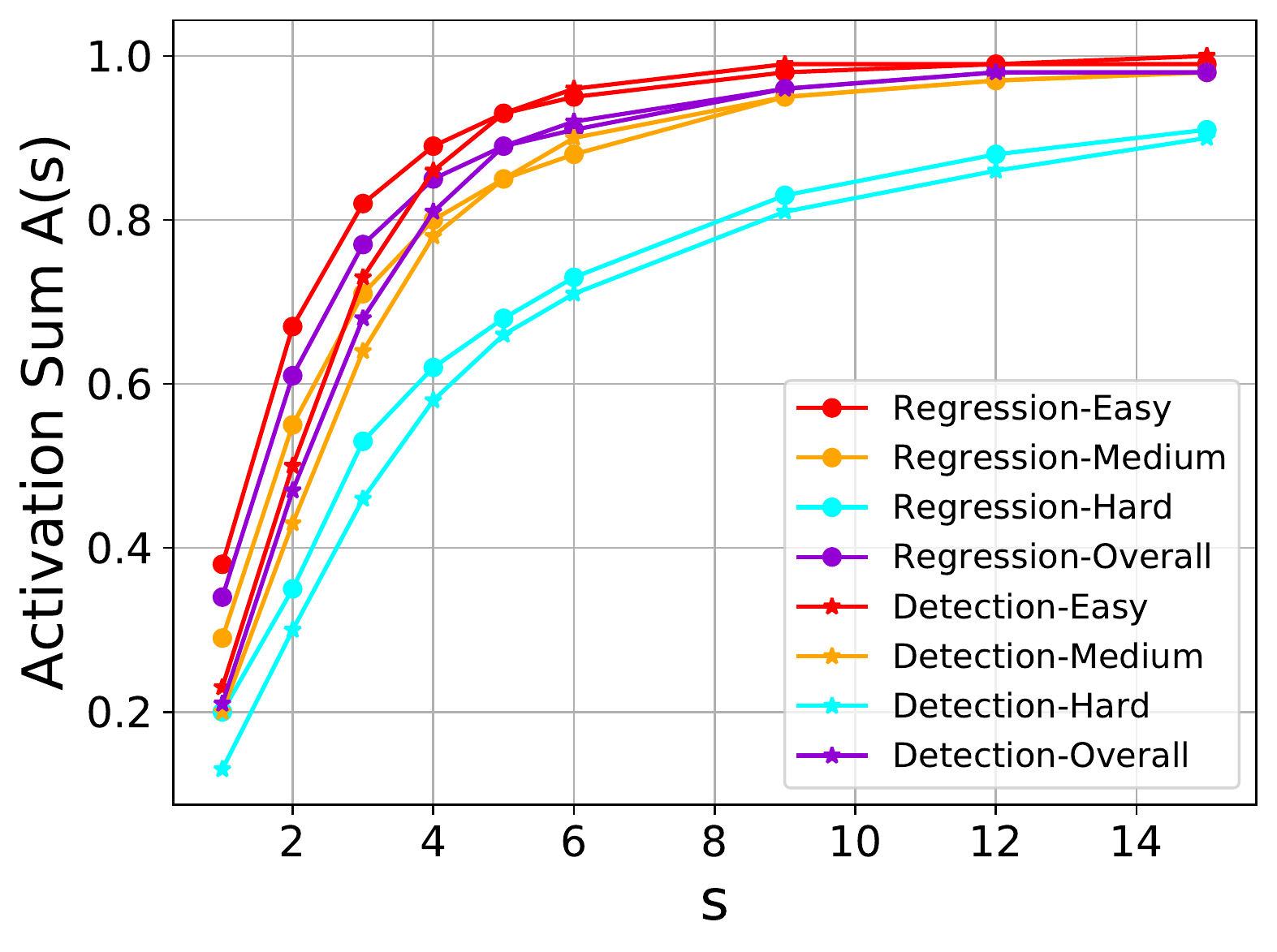}
    \caption{Activation Sum $A(s)$ of both detection and regression method on easy, medium, hard, and overall cases.}
    \label{fig:detailed_as}
\end{figure}

We provide an extended plot of Fig.~5(a) in the main paper, which includes both detection and regression on easy, medium, and hard cases in Fig.~\ref{fig:detailed_as}.

\section{Pearson Chi-square value on different cases}
\label{app:pcvalue_dist_templates}
We report the detailed Pearson Chi-square value between the heatmaps of both detection and regression in different scenarios and Gaussian templates with different $\sigma$s in Tables~\ref{tab:hard_de_re},~\ref{tab:medium_de_re}, and~\ref{tab:easy_de_re}. We set the threshold of $A(s)>0.8$ to show that the support region can represent the whole heatmaps. Therefore, for easy and medium case, $s\!=\!4$, while for hard cases, $s\!=\!8$.

\begin{table}[ht]
\centering
\resizebox{0.9\columnwidth}{!}{
\begin{tabular}{|c|c|c|c|c|c|c|c|c|}
\hline
\multirow{2}{*}{\diagbox{$s$}{$\sigma$}} & \multicolumn{4}{c|}{regression} & \multicolumn{4}{c|}{detection} \\ \cline{2-9} 
                  & 2      & 3      & 4     & 5     & 2      & 3     & 4     & 5     \\ \hline
8      & -      & 4.76    &  4.12    &  3.95  &- & 3.51    &  3.31 & 3.25        \\\hline
\end{tabular}
}
\caption{Results of Pearson Chi-square test between distribution of heatmap $\bH$ and Gaussian templates $\bT$ with different window size $s$ and standard deviation $\sigma$ in hard cases. $\sigma\!=\!5$ is the most matched template for both detection and regression.}
\label{tab:hard_de_re}
\end{table}

\begin{table}[ht]
\centering
\resizebox{0.9\columnwidth}{!}{
\begin{tabular}{|c|c|c|c|c|c|c|c|c|}
\hline
\multirow{2}{*}{\diagbox{$s$}{$\sigma$}} & \multicolumn{4}{c|}{regression} & \multicolumn{4}{c|}{detection} \\ \cline{2-9} 
            & 1      & 2      & 3      & 4   & 1  & 2      & 3     & 4   \\ \hline
4      & -     &   3.25  & 3.65    &  3.73    &  -  & 1.76  & 0.76    &  0.77        \\\hline
\end{tabular}
}
\caption{Results of Pearson Chi-square test between distribution of heatmap $\bH$ and Gaussian templates $\bT$ with different window size $s$ and standard deviation $\sigma$ in medium cases. - denotes number larger than 1000. For detection, best matched $\sigma$ should be between 3 and 4; while for regression, the optimal $\sigma$ in medium case should be between 1 and 2.}
\label{tab:medium_de_re}
\end{table}

\begin{table}[ht]
\centering
\resizebox{0.95\columnwidth}{!}{
\begin{tabular}{|c|c|c|c|c|c|c|c|c|c|c|c|}
\hline
\multirow{2}{*}{\diagbox{$s$}{$\sigma$}} & \multicolumn{5}{c|}{regression} & \multicolumn{6}{c|}{detection} \\ \cline{2-12} 
  & 0.5  & 1      & 2          & 3          & 4     & 1      & 2      & 2.5     & 3     & 3.5      & 4    \\ \hline
4      & -     &  3.24  &  3.53   &  3.75    &  3.89 &   -     &   2.52  & 0.64    &  0.72    &  0.85  &0.88      \\\hline
\end{tabular}
}
\caption{Results of Pearson Chi-square test between distribution of heatmap $\bH$ and Gaussian templates $\bT$ with different window size $s$ and standard deviation $\sigma$ in easy cases. Accordingly, for detection, $\sigma$ should be approximately 2.5; while for regression, optimal $\sigma$ should be between 0.5 and 1.}
\label{tab:easy_de_re}
\end{table}

\begin{figure*}[t]
    \centering
    \includegraphics[width=0.95\textwidth]{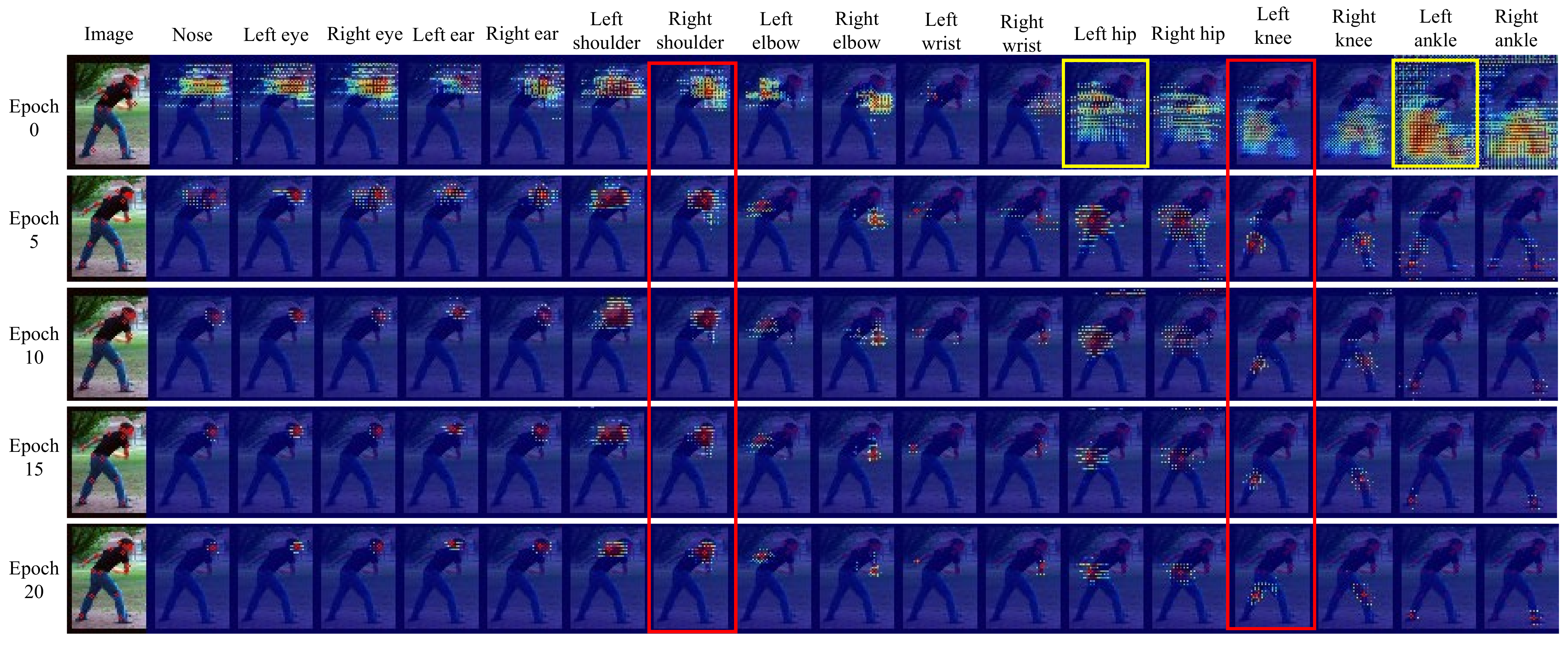}
    \caption{Visualization of training performance of integral pose regression on one single image at epoch 0, 5, 10, 15, 20 (from top to down). Corner or edge pixels of the correct quadrant are largely activated in the epoch 0 (highlighted in the yellow boxes). The further training is focused on narrowing down the activated region (shown in the red boxes).} 
    \label{fig:reg_real}
\end{figure*}

\begin{figure*}[t]
    \centering
    \includegraphics[width=0.98\textwidth]{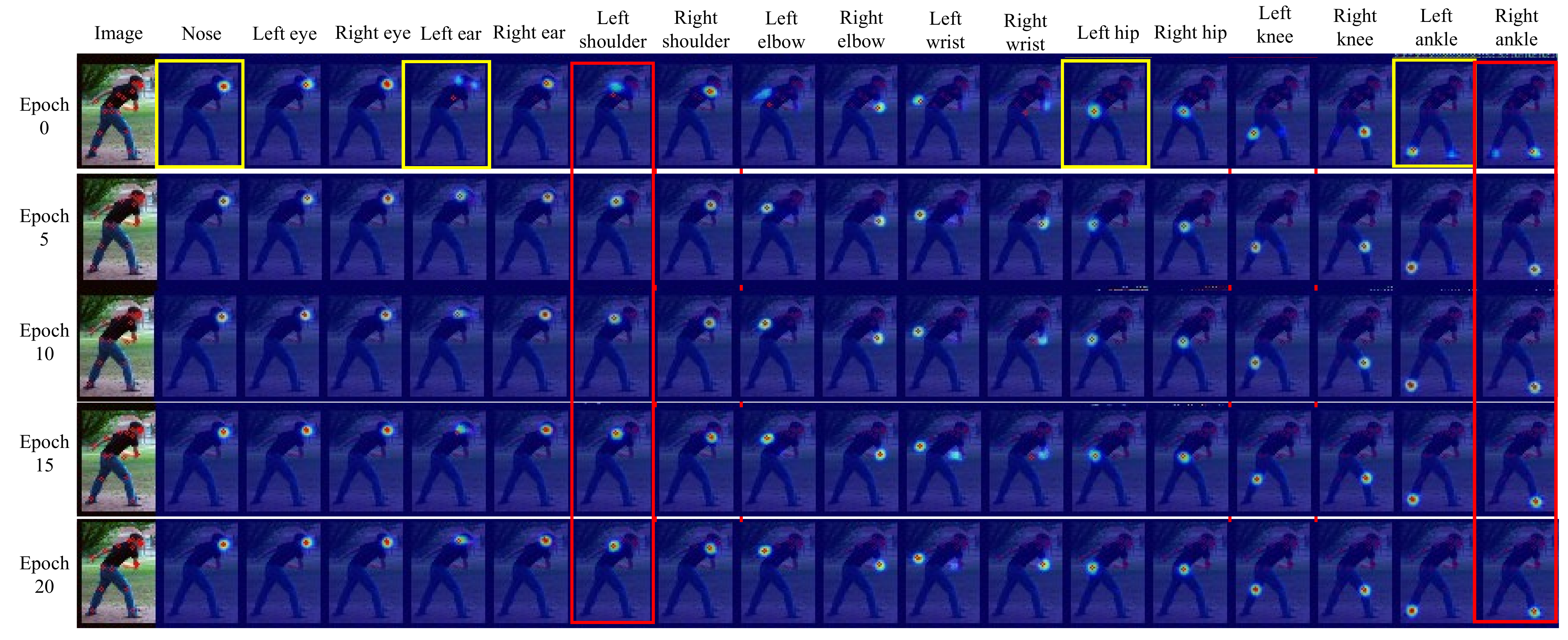}
    \caption{Visualization of training performance of detection-based methods on one single image at epoch 0, 5, 10, 15, 20 (from top to down). One epoch has already enables the network to roughly localize the joint compared with regression based method (in the yellow boxes) Further training of detection-based methods makes the predictions more stable and precise (highlighted in red boxes)}
    \label{fig:det_real}
\end{figure*}

\section{Visualizations of real applications}
\label{sec:vis_real_complete}

We show the visualization of the training performance of integral pose regression and detection-based methods on one single image in Figs.~\ref{fig:reg_real} and \ref{fig:det_real}, respectively. The further training of integral pose regression is focused on narrowing down the activated region while the further training of detection-based methods makes the predictions more stable and precise.
